\documentclass{article}

 \usepackage[main, final]{neurips_2026}

\usepackage[utf8]{inputenc} 
\usepackage[T1]{fontenc}    
\usepackage{hyperref}       
\usepackage{url}            
\usepackage{booktabs}       
\usepackage{amsfonts}       
\usepackage{nicefrac}       
\usepackage{microtype}      
\usepackage{xcolor}         
\usepackage{graphicx}
\usepackage{bm}
\usepackage{float}
\usepackage[table]{xcolor}
\usepackage{multirow}
\usepackage{amsmath}
\usepackage{color,soul}
\usepackage{amssymb}
\usepackage{mathtools}
\usepackage{enumitem}
\usepackage{algorithm}
\usepackage{algpseudocode}
\usepackage{capt-of}     
\usepackage{xcolor}
\usepackage{etoc}

\usepackage{algorithm}

\newcommand{\name}{\textbf{$\Delta$LPS}} 

\newcommand{\R}{\mathbb{R}}
\newcommand{\bc}{\mathbf{c}}
\newcommand{\bg}{\mathbf{g}}
\newcommand{\bx}{\mathbf{x}}
\newcommand{\bu}{\mathbf{u}}
\newcommand{\bw}{\mathbf{w}}
\newcommand{\be}{\mathbf{e}}
\newcommand{\bC}{\mathbf{C}}
\newcommand{\by}{\mathbf{y}}
\newcommand{\bn}{\mathbf{n}}
\newcommand{\bz}{\mathbf{z}}

\newcommand{\oA}{\mathcal{A}}
\newcommand{\oD}{\mathcal{D}}

\newcommand{\bI}{\mathbf{I}}

\newcommand{\bQ}{\mathbf{Q}}

\algrenewcommand\algorithmiccomment[1]{\hfill\textcolor{gray}{\footnotesize // #1}}

\title{Discrete-Diffusion Posterior Sampling with Langevin-Inspired Proposals}

\author{%
  Chaitanya Amballa\thanks{Equal contribution. Shared first author order determined by the toss of a fair coin.}
  \quad
  Sattwik Basu\footnotemark[1]
  \quad
  Jorge Vančo Sampedro
  \quad
  Romit Roy Choudhury
  \\
  University of Illinois Urbana-Champaign
  \\
  \texttt{\{amballa2, sattwik2, jorgev3, croy\}@illinois.edu}
}
\etocdepthtag.toc{mainpaper}
\begin{document}

\maketitle
\vspace{-1.5em}
\begin{abstract}
We study posterior sampling for inverse problems in discrete state spaces using discrete diffusion models as generative priors.
While continuous diffusion models have become widely used for inverse problems, their discrete counterparts remain comparatively underexplored.
Existing discrete posterior samplers often rely on continuous relaxations of discrete variables, Gibbs-style updates, or mechanisms specialized to particular corruption processes, which can limit scalability or generality.
We propose \name{}, a Discrete Langevin-Inspired Posterior Sampler that uses gradient information to identify promising discrete moves without leaving the discrete state space.
The resulting approach enables efficient parallel updates across all token dimensions and is agnostic to the training paradigm of the discrete diffusion prior, including masked and uniform-state diffusion.
We evaluate our method on image restoration tasks across MNIST, CIFAR, and FFHQ, spatial mapping, guided DNA-sequence generation, and monophonic music infilling, covering linear, nonlinear, and blind inverse problems.
Across these settings, we improve over recent discrete diffusion posterior samplers and are competitive with strong continuous diffusion-based inverse solvers.
Our results suggest that fully discrete, gradient-informed posterior samplers offer a scalable and general path toward solving inverse problems over discrete  representations.
  
\end{abstract}

\section{Introduction}

Inverse problems are a class of methods aimed at recovering an unknown signal $\bx$ from indirect, sparse, and noisy measurements $\by$.
The signal and measurement are typically related through a forward model
 $\by = \oA(\bx) + \bn$,
where $\oA(\cdot)$ denotes the forward operator and $\bn$ denotes measurement noise.
Given only $\by$, the goal is to infer a plausible underlying signal $\bx$ sampled from the Bayesian posterior $p(\bx | \by)$.
The central challenge is that inverse problems are often ill-posed: many distinct signals may be consistent with the same measurement.
This ambiguity makes structural priors on $\bx$ essential.
Classical signal processing has developed domain-specific priors based on sparsity, smoothness, total variation, and related regularizers \cite{EnglHankeNeubauer1996}.
These priors have enabled sophisticated optimization-based estimators, but they often underrepresent complex data and require careful tuning to balance measurement fidelity against regularization.

Recent progress has seen a paradigm shift towards using generative priors, particularly diffusion models, that have delivered substantial gains across domains \cite{nonequidiff, score, ddpm}.
Most existing posterior sampling methods, however, are designed for continuous score-based diffusion models that are naturally compatible with gradient-driven likelihood guidance \cite{score, ddpm, inversebench, dps, pigdm}.
In contrast, discrete diffusion models remain relatively underexplored for inverse problems, despite their growing role as priors over tokenized, symbolic, and quantized latent representations.
Recent work has begun to address posterior sampling in discrete state spaces \cite{sgdd, g2d2, aps}, but typically rely on one of three mechanisms: relaxing categorical variables into a continuous space, introducing augmented Gibbs-style \cite{vono2019split} updates, or specializing the sampler to a particular corruption process.
For example, G2D2 \cite{g2d2} enables likelihood guidance through a continuous relaxation, but this changes the underlying geometry on which inference is performed; SGDD \cite{sgdd} provides a principled Gibbs-based construction, but such updates can mix slowly in high-dimensional spaces; and APS \cite{aps}, while being a strong method, is specially designed around masked-diffusion models and anchoring.
These limitations motivate posterior samplers that remain fully discrete, and efficiently update all coordinates in parallel while remaining agnostic to training methodologies of the discrete diffusion prior. 
This paper aims to meet these requirements.

Our proposed algorithm is centered around the following idea:
gradients can tell us where to ``hop'' in a discrete space, even if the sampler itself cannot move continuously.
Our key idea is to appropriately utilize this gradient information (from continuous space) to choose better discrete hops, without ever leaving the discrete state space.
We formalize this intuition by using gradients to parameterize a Langevin-inspired proposal between valid discrete states, rather than optimizing continuously over soft-latents and projecting them back to tokens.
As an outcome, we meet our desired requirements.
Specifically, 
(1) we avoid a potential failure mode where a soft latent may fit the measurement well, but its projected discrete token may not decode to a similarly consistent sample,
(2) our Langevin-inspired proposal factorizes across coordinates, so transition probabilities for all dimensions can be computed and sampled in parallel, avoiding the sequential bottleneck of Gibbs-style samplers,
(3) finally, the sampler is not tied to a specific corruption process and can be used with masked, uniform-state, or other discrete diffusion priors.
Our overall contributions are summarized as follows:

\begin{figure}
    \centering
    \includegraphics[width=1.0\linewidth]{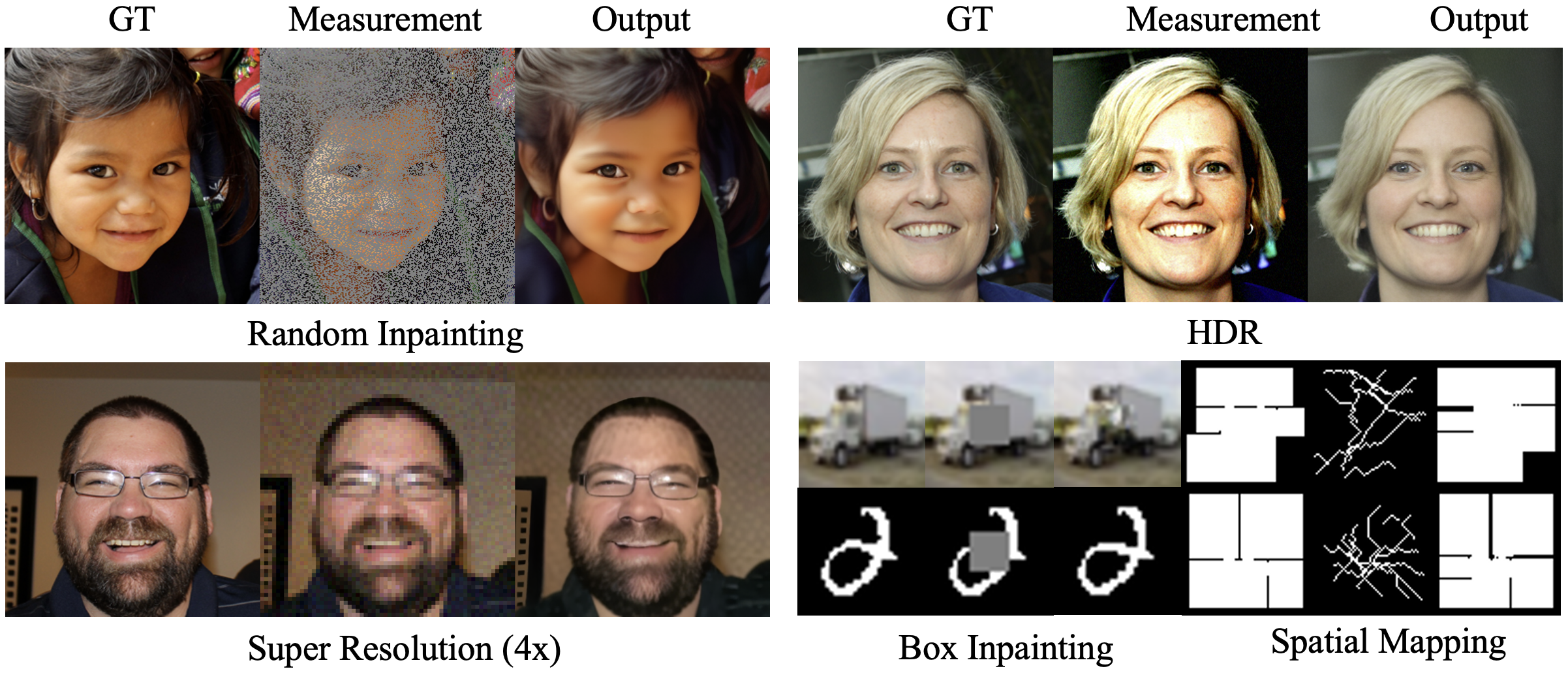}
    \vspace{-0.1in}
        \caption{\small{
  This paper introduces {\name}, a training-free approach for discrete posterior sampling using discrete diffusion priors. Shown here are results from various image restoration tasks across the MNIST, CIFAR, FFHQ datasets along with and spatial mapping. Details of other tasks such as guided DNA-sequence generation and music infilling are provided later in Sec. \ref{sec:experiments}. Project website: \url{https://discretelps.github.io}}}
    \label{fig:teaser}
\end{figure}

\begin{itemize}[leftmargin=*, nosep, itemsep=0pt, topsep=1pt, parsep=0pt, partopsep=0pt]
    \item \textbf{Fully discrete posterior sampling.}
    We introduce {\name}, a training-free discrete Langevin-inspired posterior sampler that keeps every iterate in the discrete state space, avoiding soft-token optimization and projection back to discrete tokens.

    \item \textbf{Efficient parallel updates.}
    We use gradients only to construct informed transition probabilities between valid discrete states. The resulting proposal factorizes across coordinates, enabling all token dimensions to be updated in parallel rather than through sequential Gibbs-style updates.

    \item \textbf{Prior-agnostic posterior sampler.}
    Our method only requires predictive probabilities over categorical states, making it applicable to masked, uniform-state, and other discrete diffusion priors. 
\end{itemize}

We evaluate {\name} on a broad set of inverse problems, including image restoration (Fig. \ref{fig:teaser}) on MNIST, CIFAR, and FFHQ, as well as spatial mapping, guided DNA-sequence generation, and monophonic music infilling.
These experiments cover linear, nonlinear, and blind measurement models, allowing us to test whether the sampler remains effective beyond standard observation operators.
Across these settings, our method is competitive with, or outperforms, existing discrete diffusion posterior sampling methods, and in several cases matches or exceeds strong continuous diffusion posterior sampling baselines.
The results suggest that fully discrete posterior samplers can provide a promising path toward practical and general posterior samplers over a wide class of inverse problems over discrete representations. 

\section{Preliminaries and Related Work}
\noindent \textbf{Discrete diffusion models} have recently emerged as a powerful class of generative models for \textit{discrete} data, including text \cite{llada, llada1.5, scaling1, scaling2}, code \cite{khanna2025mercury,inception_mercury_2025}, vector-quantized images \cite{mmada, lavida}, audio \cite{ yang2023diffsound}, protein, and molecular generation \cite{wang2024dplm, peptune, liang2025discrete, yin2025cfp}.
These models use a forward Markov corruption process over categorical states $\bz_0 \in \{1,\dots,K\}^{L}$ and learn a reverse denoising model.
In D3PM~\citep{d3pm}, the forward process is specified by transition matrices
$\bQ_t \in \mathbb{R}^{K \times K}$, where $\bQ_t^{ij}$ denotes the probability of transitioning from token $i$ to token $j$ at time $t$.
For each coordinate $\ell$,
$q(\bz_t[\ell]=j | \bz_{t-1}[\ell]=i) = \bQ_t^{ij},$
and the marginal corruption distribution is tractable:
$q(\bz_t | \bz_0)
    =
    \prod_{\ell=1}^{L}
    \operatorname{Cat}
    \left(
        \bz_t[\ell];
        \be_{\bz_0[\ell]}\bar{\bQ}_t
    \right),
    \bar{\bQ}_t
    =
    \bQ_1 \bQ_2 \cdots \bQ_t,$
where $\be_i$ denotes the categorical basis vector for token $i$.
The learned reverse model is typically parameterized through a denoiser
$p_\theta(\bz_0 ; \bz_t)$, which predicts clean tokens from a noisy input i.e., approximates $p(\bz_0|\bz_t)$.
This can then be combined with the analytic posterior
$q(\bz_s |\bz_t,\bz_0)$ to define a reverse transition for $s<t$:
\begin{align}
    p_\theta(\bz_s | \bz_t)
    =
    \sum_{\hat{\bz}_0}
    q(\bz_s | \bz_t,\hat{\bz}_0)
    p_\theta(\hat{\bz}_0 ; \bz_t).
\end{align}
Masked diffusion models instantiate this framework using an absorbing mask state.
For a special token $[\mathrm{M}]$, the corruption process can be written as
   $ q(\bz_t[\ell] | \bz_0[\ell])
    =
    \operatorname{Cat}
    \left(
        \bz_t[\ell];
        \alpha_t \be_{\bz_0[\ell]}
        +
        (1-\alpha_t)\be_{[\mathrm{M}]}
    \right),$

where $\alpha_t$ decreases with $t$.
MDLM-style models~\citep{mdlm} build on this absorbing process and train the denoiser with a weighted masked-token prediction objective.
A different line of work, including Duo-style uniform-state diffusion~\citep{duo}, instead uses uniform corruption,
\begin{align}
    q(\bz_t[\ell] | \bz_0[\ell])
    =
    \operatorname{Cat}
    \left(
        \bz_t[\ell];
        \alpha_t \be_{\bz_0[\ell]}
        +
        (1-\alpha_t)\bu
    \right),
    \qquad
    \bu = \frac{1}{K}\mathbf{1},
\end{align}
so corrupted variables are replaced by uniformly random tokens rather than a single absorbing mask.
Both masked and uniform-state models provide pretrained discrete priors of the form
$p_\theta(\bz_0 ; \bz_t)$, but they induce different sampling behavior and different reverse-transition structure.

\paragraph{Diffusion posterior sampling.}
Given measurements $\by$ with likelihood $p(\by | \bz_0)$, posterior sampling seeks samples from
$p(\bz_0 | \by) \propto p(\by | \bz_0)p(\bz_0)$.
In continuous score-based diffusion models \cite{score}, posterior samplers use the decomposition of the posterior score into a prior score and a likelihood score, often approximating the likelihood term through a one-step denoised estimate \citep{dps,pigdm}.
For discrete diffusion models, an analogous Euclidean score is not directly available because $\bz_t$ takes values in a finite categorical space.
Recent work has therefore explored alternative mechanisms for discrete posterior sampling.
G2D2~\citep{g2d2} addresses inverse problems with discrete diffusion priors by optimizing a variational categorical distribution using continuous relaxations, together with a star-shaped noising process designed to reduce irreversible token commitments.
SGDD~\citep{sgdd} proposes a plug-and-play split Gibbs sampler that alternates between a likelihood-side sampling step and a diffusion-prior denoising step.
APS~\citep{aps} focuses on masked diffusion foundation models and introduces quantized expectation for gradient-like guidance, together with anchored remasking to adaptively control which tokens are committed during inference.
These methods show the promise of discrete diffusion priors for inverse problems, but typically rely on continuous relaxations, Gibbs-style updates, or mechanisms specialized to a particular discrete prior.
In contrast, our goal is to construct a prior-agnostic discrete posterior sampler that uses the pretrained denoiser $p_\theta(\bz_0 ;\bz_t)$ while enabling fully parallel updates across token coordinates.
Next, we first formalize the problem in a Bayesian setting and introduce our discrete sampler.

\section{Method}

\noindent \textbf{Formulation.}
We formulate the inverse problem directly in the discrete latent space of an encoder--decoder generative model, such as a VQ-VAE \cite{oord2017vqvae}.
Let $\bz_0\in\{1,\ldots,K\}^L$ denote a clean latent sequence of $L$ discrete tokens, where $K$ is the codebook size.
The decoder $\oD$ maps this discrete latent sequence to the signal domain, giving $\bx = \oD(\bz_0) \in \mathbb{R}^n$.
The measurements are generated as $\by = \oA(\oD(\bz_0)) + \bn$, where $\oA:\mathbb{R}^n\rightarrow\mathbb{R}^m$ is the forward operator, $\by\in\mathbb{R}^m$, and $\bn\sim\mathcal{N}(\boldsymbol{0},\sigma_y^2\bI_m)$.
For inverse problems in the signals native representation (e.g., pixel space for images), $\oD$ may be interpreted as an identity operator with $\bx_t = \bz_t$.

We assume access to a pretrained discrete diffusion prior $p_\theta(\bz_0 ; \bz_t) \approx p(\bz_0 |\bz_t)$, where $\bz_t$ denotes the current noisy latent state at diffusion time $t\in[0,T]$.
Let $s<t$ denote the next lower-noise time.
\textbf{Our goal} is to sample from the posterior transition $p(\bz_s | \bz_t,\by)$.
We introduce the clean latent state $\bz_0$ as an intermediate variable between $\bz_t$ and $\bz_s$, following DAPS \cite{daps}:
\begin{align}
    p(\bz_s | \bz_t,\by)
    &=
    \sum_{\bz_0}
    p(\bz_s | \bz_t,\by,\bz_0)\,
    p(\bz_0 | \bz_t,\by) 
    \approx
    \sum_{\bz_0}
    p(\bz_s | \bz_0)\,
    p(\bz_0 | \bz_t,\by).
\end{align}
This approximation uses $\bz_0$ as the clean intermediate state: we first infer a measurement-consistent clean latent and then renoise it to time $s$.
The clean-latent posterior can be simplified as:
\begin{align}
    p(\bz_0 | \bz_t,\by)
    &=
    \frac{
        p(\by | \bz_0,\bz_t)\,
        p(\bz_0 | \bz_t)
    }{
        p(\by | \bz_t)
    } 
    \propto
    p(\by | \bz_0)\,
    p(\bz_0 | \bz_t).
    \label{eqn:daps_step_1}
\end{align}
Here, we use the conditional independence assumption that, given the clean latent $\bz_0$, the measurement $\by$ is independent of the noisy latent $\bz_t$.
This leads to the core sampling strategy.
(1) First, we sample a clean latent $\bz_0$ that is both likely under the discrete diffusion prior and consistent with the measurement: $\bz_0 \sim p(\bz_0 | \bz_t,\by) \propto p(\by | \bz_0)\, p(\bz_0 | \bz_t)$.
Unlike in continuous diffusion models, we can directly approximate $p(\bz_0 | \bz_t)$ using the discrete diffusion prior $p(\bz_0 ; \bz_t)$.
(2) Then, we renoise this clean latent to the next diffusion time $s$, i.e., $\bz_s \sim p(\bz_s | \bz_0)$.

\noindent \textbf{Discrete Langevin-inspired clean latent sampling.}
The key step is to sample the clean latent posterior
$p(\bz_0|\bz_t,\by)$ appearing in Eq.~\ref{eqn:daps_step_1}.
To motivate our sampler, first consider how the clean-latent sampling problem would be solved if $\bz_0$ were continuous.
We define the log-posterior potential
\begin{align}
    U(\bz_0;\bz_t,\by)
    =
    \log p(\by|\bz_0)
    +
    \log p_\theta(\bz_0;\bz_t),
    \label{eqn:potential}
\end{align}
whose high-probability regions correspond to clean latents that are both measurement-consistent and likely under the discrete diffusion prior.
A Langevin \cite{rrt, mala} update would move the current sample in the direction of increasing $U$, while injecting Gaussian noise to maintain stochasticity:
\begin{align}
    \bz_0'
    :=
    \bz_0
    +
    \eta \nabla_{\bz_0} U(\bz_0;\bz_t,\by)
    +
    \sqrt{2\eta}\boldsymbol{\epsilon},
    \qquad
    \boldsymbol{\epsilon}\sim\mathcal{N}(\boldsymbol{0},\bI),
\end{align}
where $\eta>0$ is a step size.
Equivalently, this corresponds to sampling $\bz_0'$ from the Gaussian proposal
\begin{align}
    q_{\mathrm{cont}}(\bz_0'|\bz_0;\bz_t,\by)
    &=
    \mathcal{N}
    \left(
        \bz_0';
        \bz_0+\eta\nabla_{\bz_0}U(\bz_0;\bz_t,\by),
        2\eta\bI
    \right) \nonumber \\
    &\propto
    \exp
    \left(
        -\frac{1}{4\eta}
        \left\|
            \bz_0'
            -
            \bz_0
            -
            \eta\nabla_{\bz_0}U(\bz_0;\bz_t,\by)
        \right\|_2^2
    \right).
\end{align}
Thus, the continuous proposal assigns high probability to states near the gradient-shifted point $\bz_0+\eta\nabla_{\bz_0}U(\bz_0;\bz_t,\by)$.
With a Metropolis--Hastings \cite{robertcasella_mcmc, speagle_mcmc} correction, this gives the standard Metropolis-Adjusted Langevin Algorithm \cite{mala, robertcasella_mcmc} proposal; without the correction, it gives the usual unadjusted Langevin sampler.

In our setting, $\bz_0$ is a discrete token sequence, so we cannot directly add a gradient vector to it and obtain another valid latent.
However, by borrowing from the discrete MCMC literature \cite{oops, dlp}, we can preserve the same principle without leaving the discrete state space.
Instead of using the Langevin update to produce a continuous-valued proposal, we use the Langevin Gaussian score to assign probabilities to valid discrete candidates $\bz_0'\in\{1,\ldots,K\}^L$:
\begin{align}
    q(\bz_0'|\bz_0;\bz_t,\by)
    &\propto
    \exp
    \left(
        -\frac{1}{4\eta}
        \left\|
            \bz_0'
            -
            \bz_0
            -
            \eta\nabla_{\bz_0}U(\bz_0;\bz_t,\by)
        \right\|_2^2
    \right).
\end{align}
This is the key continuous-to-discrete conversion: the Gaussian proposal is not sampled and rounded.
Rather, this quantity is evaluated only on valid discrete token sequences and then normalized only over that specific finite set.
Expanding the exponent
leads to a final term that does not depend on the proposed state $\bz_0'$, so it is absorbed into the normalizing constant.
Therefore, the discrete proposal can be written as
\begin{align}
    q(\bz_0'|\bz_0;\bz_t,\by)
    &\propto
    \exp
    \left(
        \frac{1}{2}\nabla_{\bz_0}U(\bz_0;\bz_t,\by)^\top(\bz_0'-\bz_0)
        -
        \frac{1}{4\eta}
        \|\bz_0'-\bz_0\|_2^2
    \right).
\end{align}

This form has a useful energy-ratio interpretation.
Using a first-order Taylor approximation around the current state $\bz_0$,
  $  U(\bz_0';\bz_t,\by)
    \approx
    U(\bz_0;\bz_t,\by)
    +
    \nabla_{\bz_0}U(\bz_0;\bz_t,\by)^\top
    \left(
        \bz_0'-\bz_0
    \right)$.
Substituting this approximation into the proposal, utilizing Eqn. \ref{eqn:potential} and simplifying the expression resultin, gives
\begin{align}
    q(\bz_0'|\bz_0;\bz_t,\by)
    &\propto
    \exp
    \left(
        \frac{1}{2}
        \left[
            U(\bz_0';\bz_t,\by)
            -
            U(\bz_0;\bz_t,\by)
        \right]
        -
        \frac{1}{4\eta}
        \|\bz_0'-\bz_0\|_2^2
    \right).\\
    &\propto
        \left[
        \frac{
            p(\by|\bz_0')
        }{
            p(\by|\bz_0)
        }
    \right]^{1/2}
    \left[
        \frac{
            p_\theta(\bz_0';\bz_t)
        }{
            p_\theta(\bz_0;\bz_t)
        }
    \right]^{1/2}
    \exp
    \left(
        -
        \frac{1}{4\eta}
        \|\bz_0'-\bz_0\|_2^2
    \right).
\end{align}

Thus, the proposal $q(\bz_0'|\bz_0;\bz_t,\by)$ favors candidates that improve the measurement likelihood, improve the diffusion-prior probability, and remain close to the current clean latent.

The same expression also yields a factorized categorical (see App. \ref{app:dlp_derivation}).
Since the exponent decomposes across token positions, the proposal factorizes as
$q(\bz_0'|\bz_0,\bz_t,\by) = \prod_{\ell=1}^L q_\ell(\bz_0'[\ell]|\bz_0,\bz_t,\by)$.
Each coordinate-wise proposal is a categorical distribution over the $K$ valid token values:
\begin{align}
    q_\ell(\bz_0'[\ell]|\bz_0;\bz_t,\by)
    = \mathrm{Cat} \left[\bz_0'[\ell]; \sigma \left(\frac{1}{2}\nabla_{\bz_0[\ell]}U(\bz_0;\bz_t,\by)
            \left(
                k-\bz_0[\ell]
            \right)
            -
            \frac{
                \left(
                    k-\bz_0[\ell]
                \right)^2
            }{
                4\eta
            }
        \right)\right].
        \label{eqn:cat_proposal_main}
\end{align}

The Softmax $\sigma$ normalizes these $K$ candidate scores at each coordinate.
This gives a fully discrete Langevin-inspired transition.
The gradient $\nabla_{\bz_0}U(\bz_0;\bz_t,\by)$ is computed from the full latent sequence, so each coordinate update is informed by the global measurement likelihood and the diffusion prior.
However, once the logits are formed, all $L$ categorical distributions can be sampled independently and in parallel:
   $ \bz_0'[\ell]
    \sim
    q_\ell(\cdot|\bz_0,\bz_t,\by),
    ~~
    \ell=1,\ldots,L.$
Therefore, the sampler uses posterior gradients to guide transitions toward high-probability clean latents, while ensuring that every proposed state remains a valid discrete token sequence.

\subsection{Generalizing across Inverse Problems}
\vspace{-0.05in}
The proposed sampler only requires access to the likelihood-gradient component of the posterior potential
$U(\bz_0;\bz_t,\by)=\log p(\by|\bz_0)+\log p_\theta(\bz_0;\bz_t)$.
This makes the method applicable to a broad class of inverse problems.
For known linear or nonlinear forward operators under the Gaussian noise model, the likelihood is explicit, and its contribution to the sampler is obtained by differentiating the measurement residual:
   $ \nabla_{\bz_0}\log p(\by|\bz_0)
    =
    -
    \frac{1}{2\sigma_y^2}
    \nabla_{\bz_0}
    \left\|
        \by-\oA(\oD(\bz_0))
    \right\|_2^2.$
When $\oA$ and $\oD$ are differentiable, this gradient can be computed directly by automatic differentiation.
This covers standard linear inverse problems such as inpainting, super-resolution, and deblurring, as well as nonlinear inverse problems with known differentiable forward operators.

For fully blind or non-differentiable inverse problems, the likelihood $p(\by|\bz_0)$ may be unavailable or difficult to differentiate.
In such settings, we draw on recent advances in blind posterior sampling \cite{coguide, cldps, e&e} and replace the likelihood gradient with a learned contrastive \cite{jaiswal2021surveycontrastive, supcon, infonce} surrogate.
Specifically, we can leverage two encoders $f_\varphi$ and $g_\psi$ to map decoded clean latents and measurements into a shared smooth embedding space, where compatible pairs are close and mismatched pairs are separated.
The likelihood-gradient term is then approximated as
    $\nabla_{\bz_0}\log p(\by|\bz_0)
    \approx
    \frac{1}{\tau}
    \nabla_{\bz_0}
    \left\langle
        f_\varphi(\oD(\bz_0)),
        g_\psi(\by)
    \right\rangle,$
where $\tau>0$ is a temperature parameter.
This surrogate provides a smooth guidance signal even when the true forward process is unknown, non-differentiable, or only implicitly specified.

Therefore, the same discrete Langevin-inspired sampler applies across known, nonlinear, and blind inverse problems.
Crucially, changing the inverse problem only changes the likelihood term used inside $U(\bz_0;\bz_t,\by)$; the discrete posterior proposal and the parallel categorical update remain unchanged.

\subsection{Adam Preconditioning and Renoising}
\vspace{-0.05in}
In high-dimensional discrete latent spaces, the raw posterior gradient
    $\nabla_{\bz_0} U(\bz_0;\bz_t,\by)$
can be poorly scaled across coordinates. 
Some latent coordinates may receive large, noisy gradients, while others receive gradients that are consistently small but informative. Directly using the raw gradient in the discrete Langevin proposal can therefore lead to unstable jumps in some coordinates and slow progress in others.

To address this, we replace the raw gradient with an Adam \cite{adam} preconditioned direction.
Thus, Adam does not change the structure of the sampler; it only replaces the raw posterior gradient with an adaptively rescaled update direction.
Intuitively, Adam acts as a diagonal preconditioner that normalizes the gradient using its recent first- and second-moment statistics. This makes the proposal less sensitive to coordinate-wise differences in gradient scale. Large, noisy directions are damped, while consistently useful directions are preserved. This is especially helpful in VQ or discrete latent spaces, where different token coordinates can have very different effects on the decoded image and measurement loss. In practice, this improves mixing and convergence by producing more balanced discrete proposals across the latent grid.

\noindent \textbf{Overall algorithm.} This gives a complete transition from $\bz_t$ to $\bz_s$: the sampler refines a clean latent estimate $\bz_0$ using the discrete langevin-inspired proposal mechanics, then returns to the diffusion trajectory through the discrete renoising kernel.
This process is repeated for all diffusion time steps $t\rightarrow0$.
Our complete algorithmic pipeline is explained in Algo. \ref{alg:discrete_langevin_sampler}.

\section{Experiments}
\label{sec:experiments}
\vspace{-0.1in}

We now describe our experiments that include 1) image restoration, 2) spatial mapping, 3) music infilling, and 4) guided DNA-sequence generation. 

\noindent\textbf{Image Baselines.}
We evaluate the performance of \name\ against other discrete diffusion-based solvers such as SGDD \cite{sgdd}, G2D2 \cite{g2d2}, and APS~\cite{aps}.
To demonstrate robustness, we also compare with other continuous diffusion-based baselines such as DPS~\cite{dps}, PSLD~\cite{psld}, DAPS~\cite{daps}, and ReSample \cite{resample}.

\noindent\textbf{Datasets.} We evaluate \name\ on a diverse suite of linear and non-linear inverse problems across three image datasets: FFHQ \cite{ffhq}, MNIST \cite{mnist}, and CIFAR-10 \cite{cifar}.
FFHQ consists of high-resolution images of human faces, while MNIST and CIFAR-10 are standard evaluation datasets containing handwritten digits and natural object images, respectively.
We perform inference on FFHQ in the latent space of a pretrained encoder-decoder architecture coupled with a pretrained prior model, using images of resolution \(256 \times 256\). 
The first 100 images of the FFHQ val set are used for evaluation.
For MNIST and CIFAR-10, we operate directly in pixel space on \(32 \times 32\) images; MNIST is represented as a single-channel binary image dataset with vocabulary size 2, whereas CIFAR-10 consists of three-channel color images with vocabulary size 256. 
We train both MDLM-style and Duo-style priors on MNIST and CIFAR-10.

\noindent\textbf{Tasks.} 
$\blacksquare$ \textbf{FFHQ}: We consider random inpainting, Gaussian deblurring, super resolution (4$\times$), high dynamic range (HDR), and motion deblur.
For each of the tasks, we choose the measurement operator and the noise parameters following the standard settings in the baselines \cite{dps, g2d2}.
For each image, we apply the measurement operator to obtain the observed measurement $\by$, and use \name{} to sample from the posterior $p(\bz_0 | \by)$.
The final image is then reconstructed by decoding the sampled tokens $\bz_0$ back to pixel space using the pre-trained decoder.
$\blacksquare$ \textbf{MNIST}: 
In addition to the standard random pixel inpainting, we also consider the box-inpainting setting, where we try to recover a specific area of the image that is masked. 
Furthermore, we also evaluate on binary operators such as XOR ($\by = p_1 \oplus p_2$) and AND ($\by = p_1 \wedge p_2$) for MNIST, i.e., given two pixel measurements $p_1, p_2 \in \{0,1\}$ drawn from positions $i$ and $j$ in the image.
where $\oplus$ and $\wedge$ denote the element-wise exclusive-or and conjunction over the token vocabulary, respectively.
We randomly select $\nu$ pairs of pixels to obtain the final measurement set.
$\blacksquare$ \textbf{CIFAR-10}: Similar to MNIST, we consider both the inpainting tasks to evaluate on CIFAR-10. 
The discrete binary operators discussed above do not apply here.
To study robustness to ill-posedness, i.e., how more corruption leads to more valid solutions, we construct three tiers of difficulty: \emph{Easy}, \emph{Medium}, and \emph{Hard}, by progressively increasing the measurement corruption.
For instance, in the random inpainting, we increase the percentage of masked pixels and higher noise levels as the difficulty increases.


\noindent\textbf{Metrics.}
For FFHQ, we report PSNR (dB) and LPIPS~\cite{lpips}, which together capture both fidelity and perceptual quality.
PSNR measures the pixel-wise fidelity of the reconstructed image to the ground truth, while LPIPS captures perceptual similarity by comparing features extracted from a pre-trained network.
For MNIST, along with PSNR, we also report per-token (pixel) accuracy (\%). 
We report PSNR and SSIM, which together capture both fidelity and structural similarity for CIFAR-10.
All the metrics and baselines are evaluated on their standard sizes.

\subsection{Quantitative Results.}
\vspace{-0.05in}
Table \ref{tab:ffhq} reports comparisons between \name{} and baselines across five inverse problems on FFHQ.
We report both mean and standard deviation across the validation dataset and across 5 different random seeds.
\name{}'s performance is comparable to, or better than, both continuous and discrete baselines.
In inpainting, Gaussian deblur, and super resolution (4$\times$), our method outperforms all the discrete baselines in LPIPS and is on par in terms of PSNR.
PSNR drops in other tasks in HDR and Motion Deblur, but remains strong in LPIPS, suggesting that the reconstructions are perceptually more faithful to the original images.
Gains in APS can be partially attributed to the anchoring mechanism to control the denoising process.
Our vanilla \name\ doesn't incorporate any prior-specific techniques.
We expect our results to be only stronger with any such additional controls.
\footnote{As of writing this manuscript, we do not have access to APS codebase, so we report the numbers from their paper.
We have tried to reproduce their results but have not been able to match their performance, which may be due to differences in implementation details or hyperparameters.}


\begin{table}[H]
\centering
\footnotesize
\setlength{\tabcolsep}{3pt}
\caption{FFHQ inv. prob: PSNR (dB)$\uparrow$ and LPIPS$\downarrow$ with std.\ shown as subscript. Best shown in \textbf{bold}.}
\label{tab:ffhq}

\resizebox{\columnwidth}{!}{
\begin{tabular}{llcccccccccc}
\toprule
\textbf{Type} & \textbf{Method}
    & \multicolumn{2}{c}{\textbf{Inpainting}}
    & \multicolumn{2}{c}{\textbf{Deblur}}
    & \multicolumn{2}{c}{\textbf{SR4$\times$}}
    & \multicolumn{2}{c}{\textbf{HDR}}
    & \multicolumn{2}{c}{\textbf{Motion Deblur}} \\

\cmidrule(lr){3-4} \cmidrule(lr){5-6}
\cmidrule(lr){7-8} \cmidrule(lr){9-10}
\cmidrule(lr){11-12}

 &  & PSNR$\uparrow$ & LPIPS$\downarrow$
    & PSNR$\uparrow$ & LPIPS$\downarrow$
    & PSNR$\uparrow$ & LPIPS$\downarrow$
    & PSNR$\uparrow$ & LPIPS$\downarrow$
    & PSNR$\uparrow$ & LPIPS$\downarrow$ \\

\midrule


\rowcolor{gray!15}
 & DPS
    & 25.46 & 0.203   & 25.87 & 0.219
    & 25.86 & 0.269   & 22.73 & 0.264
    & 24.52 & 0.246 \\

\rowcolor{gray!15}
 & PSLD
    & 30.31 & 0.221   & 27.37 & 0.304
    & 27.62 & 0.276   & --- & ---
    & 22.31 & 0.336 \\

\rowcolor{gray!15}
 & DAPS
    & 31.12 & 0.098   & 29.19 & 0.165
    & 29.07 & 0.177   & 27.12 & 0.162
    & 29.66 & 0.157 \\

\rowcolor{gray!15}
 & DDRM
    & --- & ---       & 24.93 & 0.239
    & 26.58 & 0.282   & --- & ---
    & --- & --- \\

\rowcolor{gray!15}
 & DiffPIR
    & --- & ---       & 27.36 & 0.236
    & 26.64 & 0.260   & --- & ---
    & 26.57 & 0.255 \\

\rowcolor{gray!15}
 & ReSample
    & 29.61 & 0.140    & 25.69 & 0.329
    & 22.98 & 0.507   & 25.65 & 0.182
    & 27.41 & 0.198 \\

\rowcolor{gray!15}
\multirow{-7}{*}{Continuous} & LatentDAPS
    & 30.71 & 0.141   & 27.93 & 0.234
    & 27.48 & 0.182   & 25.94 & 0.223
    & 27.00 & 0.283 \\

\addlinespace[2pt]
\midrule


\rowcolor{orange!15}
 & SGDD
    & --- & ---       & --- & ---
    & 25.85 & 0.288   & --- & ---
    & --- & --- \\

\rowcolor{orange!15}
 & G2D2
    & --- & ---       & 26.91 & 0.280
    & 27.29 & 0.265   & --- & ---
    & --- & --- \\

\rowcolor{orange!15}
 & APS
    & 27.38 & 0.304   & 27.90 & 0.276
    & 27.50 & 0.234   & 23.89 & 0.282
    & 26.58 & 0.317 \\

\rowcolor{blue!10}
\multirow{-4}{*}{Discrete} & {\name}
    & 27.29 & \textbf{0.172}
    & 26.15 & \textbf{0.276}
    & 25.93 & \textbf{0.230}
    & 20.97 & \textbf{0.241}
    & 21.00 & \textbf{0.260} \\

\bottomrule
\end{tabular}
}
\end{table}

Tables~\ref{tab:table2-mnist} and~\ref{tab:table2-cifar} report results across various difficulty levels on MNIST and CIFAR-10.
To demonstrate \name{} is prior-agnostic, we report results using both the MDLM and \textsc{DUO} priors, which we train with different noise schedules and training objectives. 
For SGDD, we report two variants: SGDD \emph{fair} that matches our method's budget, and SGDD \emph{full}, which has $6400 \times$ more function-evaluation budget than \name\ and $100\times$ than SGDD \emph{fair}. 
On MNIST, both MDLM and \textsc{DUO} priors consistently match or surpass \textsc{SGDD} (\emph{full} and \emph{fair}) and G2D2 across all four operators.
SGDD \emph{full} is the only competitive baseline in a few instances, likely aided by its substantially larger evaluation budget. 
Nevertheless, we outperform \textsc{SGDD} in all the other settings and consistently surpass SGDD \emph{fair} across all instances.
Of course, as the difficulty increases, the performance of all the methods degrades, but \name{} maintains a consistent advantage over the baselines, demonstrating its robustness to ill-posedness.
Similarly, on CIFAR-10, \name{} (\textsc{DUO}) leads across both inpainting and box-inpainting at all difficulty levels, with particularly large gains in the hard regime. 

\begin{table}[hbt]
    \centering
    \footnotesize
    \setlength{\tabcolsep}{3pt}
    \caption{Performance across inverse problems on MNIST. Token accuracy (\%) and PSNR (dB). Both mean and std.\ dev.\ are reported. The best runs are shown in \textbf{bold}.}
    \label{tab:table2-mnist}

    \resizebox{\columnwidth}{!}{%
\begin{tabular}{llcccccccc}
\toprule
\textbf{Difficulty} & \textbf{Method}
    & \multicolumn{2}{c}{\textbf{Inpaint}}
    & \multicolumn{2}{c}{\textbf{Box}}
    & \multicolumn{2}{c}{\textbf{XOR}}
    & \multicolumn{2}{c}{\textbf{AND}} \\
\cmidrule(lr){3-4} \cmidrule(lr){5-6}
\cmidrule(lr){7-8} \cmidrule(lr){9-10}
 &  & Acc.$\uparrow$ & PSNR$\uparrow$
    & Acc.$\uparrow$ & PSNR$\uparrow$
    & Acc.$\uparrow$ & PSNR$\uparrow$
    & Acc.$\uparrow$ & PSNR$\uparrow$ \\
\midrule

\rowcolor{gray!15}
 & \textsc{SGDD} (full)
    & $99.16_{\pm 0.31}$ & $21.08_{\pm 1.83}$
    & $\bm{97.64_{\pm 1.28}}$ & $\bm{17.19_{\pm 3.22}}$
    & $\bm{99.41_{\pm 0.31}}$ & $\bm{22.90_{\pm 2.47}}$
    & $98.79_{\pm 1.01}$ & $20.95_{\pm 4.29}$ \\

\rowcolor{gray!15}
 & \textsc{SGDD} (fair)
    & $95.19_{\pm 2.29}$ & $14.02_{\pm 3.21}$
    & $93.53_{\pm 1.73}$ & $12.13_{\pm 1.66}$
    & $94.38_{\pm 2.53}$ & $13.18_{\pm 2.74}$
    & $90.04_{\pm 3.28}$ & $10.29_{\pm 1.60}$ \\

\rowcolor{gray!15}
 & G2D2
    & $97.75_{\pm 0.76}$ & $16.71_{\pm 1.42}$
    & $95.94_{\pm 0.61}$ & $13.96_{\pm 0.63}$
    & $98.30_{\pm 0.61}$ & $17.99_{\pm 1.70}$
    & $98.06_{\pm 0.93}$ & $17.70_{\pm 2.37}$ \\

\rowcolor{gray!15}
 & {\name} (MDLM)
    & $98.90_{\pm 0.41}$ & $20.01_{\pm 2.16}$
    & $97.38_{\pm 1.37}$ & $16.43_{\pm 2.41}$
    & $98.91_{\pm 0.51}$ & $20.27_{\pm 2.66}$
    & $98.54_{\pm 0.87}$ & $19.62_{\pm 3.89}$ \\

\rowcolor{gray!15}
\multirow{-5}{*}{Easy} & {\name} (\textsc{DUO})
    & $\bm{99.27_{\pm 0.31}}$ & $\bm{21.72_{\pm 1.79}}$
    & $97.48_{\pm 1.26}$ & $16.56_{\pm 2.33}$
    & $97.91_{\pm 0.48}$ & $16.91_{\pm 1.04}$
    & $\bm{99.04_{\pm 0.57}}$ & $\bm{21.33_{\pm 3.57}}$ \\

\addlinespace[2pt]
\midrule

\rowcolor{blue!10}
 & \textsc{SGDD} (full)
    & $\bm{97.89_{\pm 0.93}}$ & $\bm{17.21_{\pm 2.06}}$
    & $90.93_{\pm 3.01}$ & $10.71_{\pm 1.70}$
    & $97.74_{\pm 1.09}$ & $17.02_{\pm 2.27}$
    & $94.74_{\pm 3.67}$ & $14.35_{\pm 4.08}$ \\

\rowcolor{blue!10}
 & \textsc{SGDD} (fair)
    & $91.79_{\pm 2.96}$ & $11.22_{\pm 1.97}$
    & $89.98_{\pm 4.38}$ & $10.56_{\pm 2.46}$
    & $90.99_{\pm 3.90}$ & $10.98_{\pm 2.32}$
    & $88.54_{\pm 3.99}$ & $9.70_{\pm 1.67}$ \\

\rowcolor{blue!10}
 & G2D2
    & $94.27_{\pm 1.19}$ & $12.51_{\pm 0.86}$
    & $89.73_{\pm 1.16}$ & $9.91_{\pm 0.50}$
    & $94.88_{\pm 1.16}$ & $13.04_{\pm 1.11}$
    & $95.37_{\pm 1.01}$ & $13.44_{\pm 0.94}$ \\

\rowcolor{blue!10}
 & {\name} (MDLM)
    & $97.01_{\pm 1.12}$ & $15.55_{\pm 1.66}$
    & $\bm{92.15_{\pm 2.85}}$ & $\bm{11.46_{\pm 2.08}}$
    & $96.52_{\pm 1.23}$ & $15.00_{\pm 2.15}$
    & $97.03_{\pm 1.04}$ & $15.62_{\pm 1.87}$ \\

\rowcolor{blue!10}
\multirow{-5}{*}{Medium} & {\name} (\textsc{DUO})
    & $97.61_{\pm 0.98}$ & $16.65_{\pm 2.05}$
    & $91.31_{\pm 2.57}$ & $10.82_{\pm 1.37}$
    & $\bm{98.34_{\pm 0.65}}$ & $\bm{18.23_{\pm 2.08}}$
    & $\bm{97.65_{\pm 0.90}}$ & $\bm{16.59_{\pm 1.61}}$ \\

\addlinespace[2pt]
\midrule

\rowcolor{orange!15}
 & \textsc{SGDD} (full)
    & $\bm{95.06_{\pm 2.53}}$ & $\bm{13.72_{\pm 2.55}}$
    & $85.78_{\pm 4.00}$ & $8.68_{\pm 1.42}$
    & $93.19_{\pm 3.14}$ & $12.23_{\pm 2.37}$
    & $91.97_{\pm 4.19}$ & $11.78_{\pm 2.95}$ \\

\rowcolor{orange!15}
 & \textsc{SGDD} (fair)
    & $89.37_{\pm 3.68}$ & $10.07_{\pm 1.86}$
    & $87.32_{\pm 4.26}$ & $9.28_{\pm 1.79}$
    & $88.73_{\pm 3.73}$ & $9.74_{\pm 1.56}$
    & $88.42_{\pm 4.67}$ & $9.84_{\pm 2.28}$ \\

\rowcolor{orange!15}
 & G2D2
    & $90.74_{\pm 1.31}$ & $10.38_{\pm 0.61}$
    & $84.47_{\pm 2.14}$ & $8.13_{\pm 0.63}$
    & $89.90_{\pm 2.00}$ & $10.05_{\pm 0.92}$
    & $92.34_{\pm 1.04}$ & $11.20_{\pm 0.58}$ \\

\rowcolor{orange!15}
 & {\name} (MDLM)
    & $94.04_{\pm 2.49}$ & $12.69_{\pm 2.08}$
    & $\bm{88.51_{\pm 4.27}}$ & $\bm{9.80_{\pm 2.05}}$
    & $94.22_{\pm 2.67}$ & $12.90_{\pm 2.23}$
    & $94.57_{\pm 1.83}$ & $12.94_{\pm 1.65}$ \\

\rowcolor{orange!15}
\multirow{-5}{*}{Hard} & {\name} (\textsc{DUO})
    & $93.77_{\pm 2.46}$ & $12.58_{\pm 2.41}$
    & $86.55_{\pm 3.69}$ & $8.90_{\pm 1.31}$
    & $\bm{94.37_{\pm 3.14}}$ & $\bm{13.22_{\pm 2.61}}$
    & $\bm{94.95_{\pm 1.74}}$ & $\bm{13.25_{\pm 1.63}}$ \\

\bottomrule
\end{tabular}
}
\vspace{-1em}
\end{table}

\subsection{Qualitative  Results.}
\begin{figure*}[hbt]
    \centering
    \includegraphics[width=1.0\linewidth]{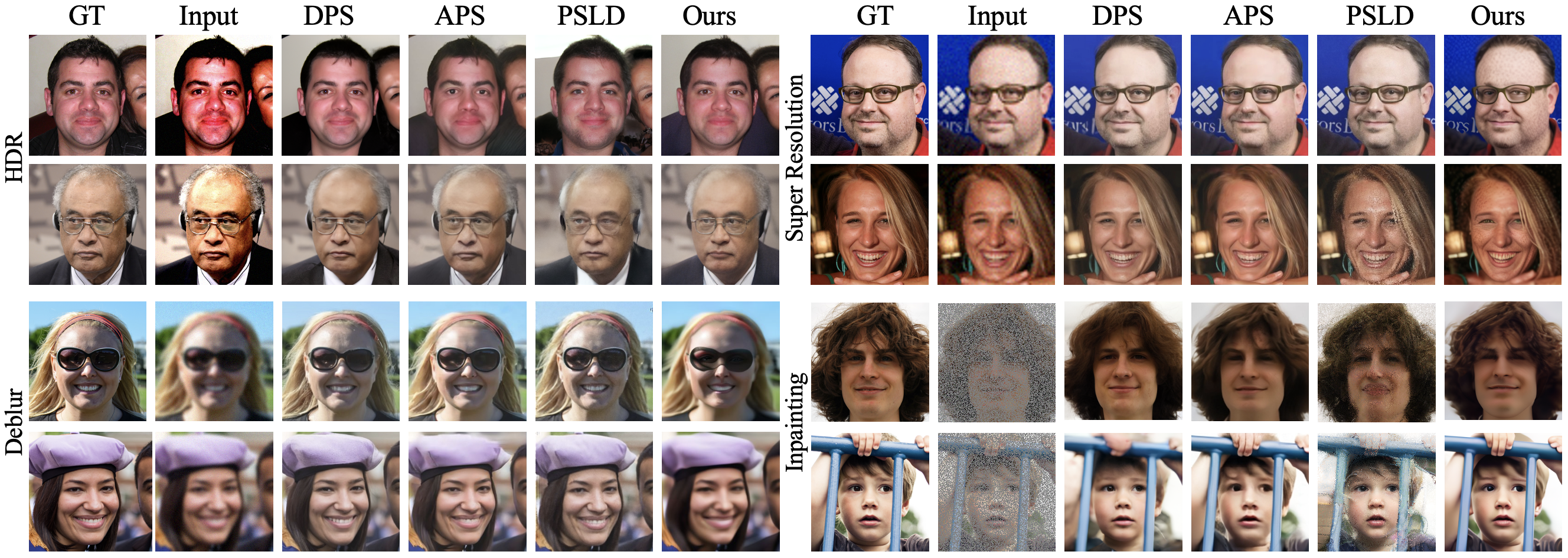}
    \vspace{-1em}
    \caption{Comparison on FFHQ across multiple inverse problems, including HDR reconstruction, $4\times$ super-resolution, Gaussian deblurring, and random inpainting. We compare \name{} against continuous diffusion-based baselines such as DPS and PSLD, as well as the discrete baseline APS. Across all tasks, reconstructions are visually comparable to those of the baselines, preserving facial details and perceptual quality even under severe corruptions.}
    \label{fig:ffhq}
\end{figure*}

Fig \ref{fig:ffhq} shows qualitative results on FFHQ compared to other baselines, both continuous and discrete.
\name{} produces reconstructions that are more visually faithful to the original images, with sharper details and fewer artifacts compared to baselines.
Even in the non-linear setting, such as HDR, \name\ remains strongly competitive to the continuous baselines and better quality than APS.
We also show qualitative results on motion blur and several tasks on MNIST in Fig \ref{fig:motion_mnist}. 
Our methods remain consistent with the ground truth under stronger corruptions, when compared to APS, and often recover plausible alternate solutions.
For instance, although the bottom right image does not exactly match the ground truth digit ``6'', our output is still a valid solution to the inverse problem.
More evaluation results and complete details on implementation can be found in Appendix \ref{app:implementation}.

\vspace{-0.5em}

\vspace{-1em}
\begin{table}[htb]
    \centering
    \footnotesize
    \setlength{\tabcolsep}{10pt}
    \caption{CIFAR inverse problems. PSNR (dB) and SSIM with std as subscript}
    \label{tab:table2-cifar}

\begin{tabular}{llcccc}
\toprule
\textbf{Diff.} & \textbf{Method}
    & \multicolumn{2}{c}{\textbf{Inpaint}}
    & \multicolumn{2}{c}{\textbf{Box}} \\
\cmidrule(lr){3-4} \cmidrule(lr){5-6}
 &  & PSNR$\uparrow$ & SSIM$\uparrow$
    & PSNR$\uparrow$ & SSIM$\uparrow$ \\
\midrule

\rowcolor{gray!15}
 & \textsc{SGDD} (full)
    & $10.20_{\pm 0.63}$ & $0.382_{\pm 0.087}$
    & $10.52_{\pm 0.76}$ & $0.398_{\pm 0.071}$ \\

\rowcolor{gray!15}
 & \textsc{SGDD} (fair)
    & \multicolumn{1}{c}{---} & \multicolumn{1}{c}{---}
    & \multicolumn{1}{c}{---} & \multicolumn{1}{c}{---} \\

\rowcolor{gray!15}
 & G2D2
    & $23.54_{\pm 2.02}$ & $0.938_{\pm 0.020}$
    & $22.45_{\pm 2.41}$ & $0.909_{\pm 0.032}$ \\

\rowcolor{gray!15}
 & {\name} (MDLM)
    & $21.79_{\pm 1.15}$ & $0.905_{\pm 0.039}$
    & $22.19_{\pm 2.31}$ & $0.908_{\pm 0.027}$ \\

\rowcolor{gray!15}
\multirow{-5}{*}{Easy} & {\name} (\textsc{DUO})
    & $\bm{28.92_{\pm 2.58}}$ & $\bm{0.978_{\pm 0.014}}$
    & $\bm{23.33_{\pm 2.92}}$ & $\bm{0.911_{\pm 0.035}}$ \\

\addlinespace[2pt]
\midrule

\rowcolor{blue!10}
 & \textsc{SGDD} (full)
    & $10.11_{\pm 0.70}$ & $0.373_{\pm 0.073}$
    & $10.44_{\pm 0.69}$ & $0.388_{\pm 0.070}$ \\

\rowcolor{blue!10}
 & \textsc{SGDD} (fair)
    & \multicolumn{1}{c}{---} & \multicolumn{1}{c}{---}
    & \multicolumn{1}{c}{---} & \multicolumn{1}{c}{---} \\

\rowcolor{blue!10}
 & G2D2
    & $18.87_{\pm 1.33}$ & $0.828_{\pm 0.041}$
    & $17.48_{\pm 1.82}$ & $0.75_{\pm 0.049}$ \\

\rowcolor{blue!10}
 & {\name} (MDLM)
    & $17.15_{\pm 1.08}$ & $0.766_{\pm 0.066}$
    & $16.68_{\pm 1.75}$ & $0.739_{\pm 0.056}$ \\

\rowcolor{blue!10}
\multirow{-5}{*}{Medium} & {\name} (\textsc{DUO})
    & $\bm{22.02_{\pm 1.62}}$ & $\bm{0.911_{\pm 0.031}}$
    & $\bm{17.95_{\pm 1.61}}$ & $\bm{0.757_{\pm 0.046}}$ \\

\addlinespace[2pt]
\midrule

\rowcolor{orange!15}
 & \textsc{SGDD} (full)
    & $9.71_{\pm 0.76}$ & $0.350_{\pm 0.083}$
    & $10.31_{\pm 0.62}$ & $0.360_{\pm 0.073}$ \\

\rowcolor{orange!15}
 & \textsc{SGDD} (fair)
    & \multicolumn{1}{c}{---} & \multicolumn{1}{c}{---}
    & \multicolumn{1}{c}{---} & \multicolumn{1}{c}{---} \\

\rowcolor{orange!15}
 & G2D2
    & $15.08_{\pm 0.92}$ & $0.651_{\pm 0.060}$
    & $\bm{14.34_{\pm 1.51}}$ & $\bm{0.581_{\pm 0.064}}$ \\

\rowcolor{orange!15}
 & {\name} (MDLM)
    & $13.58_{\pm 1.39}$ & $0.592_{\pm 0.114}$
    & $13.59_{\pm 1.30}$ & $0.561_{\pm 0.079}$ \\

\rowcolor{orange!15}
\multirow{-5}{*}{Hard} & {\name} (\textsc{DUO})
    & $\bm{15.75_{\pm 1.56}}$ & $\bm{0.706_{\pm 0.066}}$
    & $13.66_{\pm 1.13}$ & $0.560_{\pm 0.066}$ \\

\bottomrule
\end{tabular}
\vspace{-1.5em}
\end{table}

\subsection{Blind Inverse Problems.}
\begin{figure*}
    \centering
    \includegraphics[width=1\linewidth]{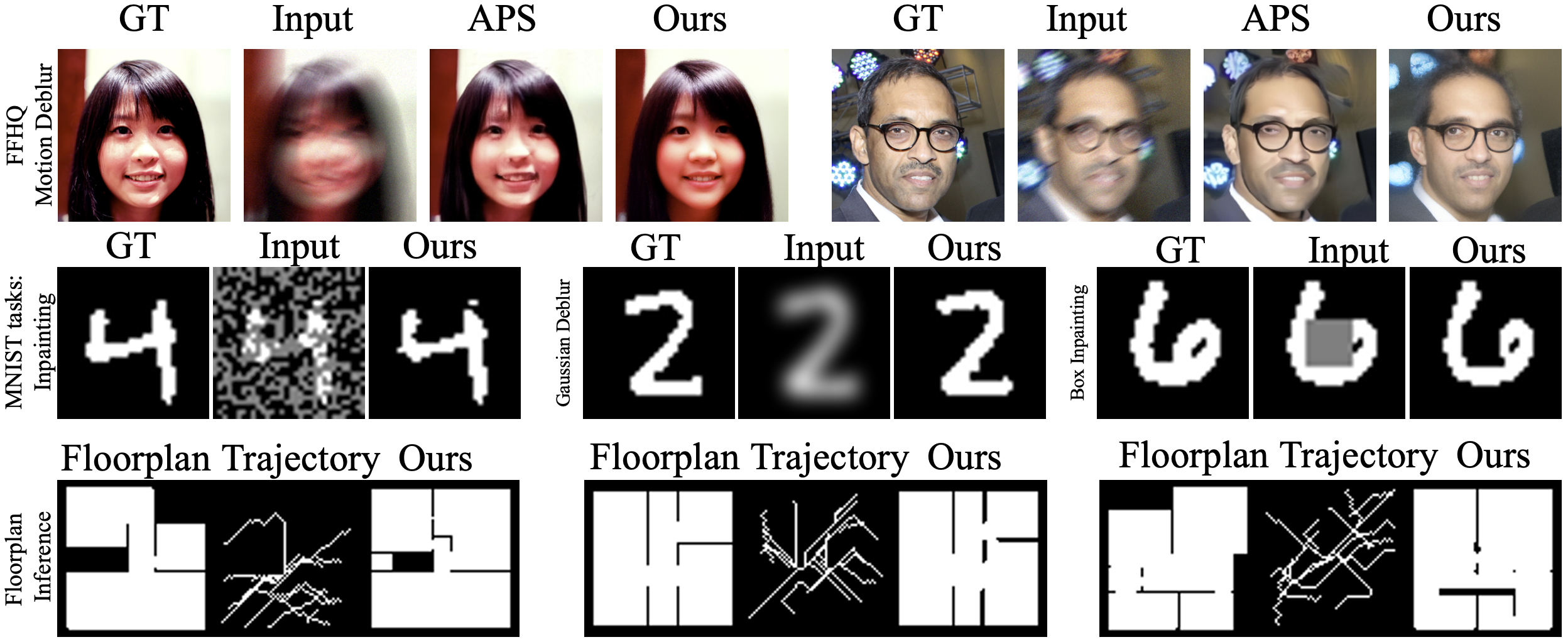}
    \vspace{-1.5em}
    \caption{\small{Qualitative results showing \name{}'s performance across various settings. On top, we show motion deblurring on FFHQ, outperforming the discrete APS baseline. In the middle, we evaluate on three different tasks on MNIST. The last row shows performance on the blind inverse setting, where the floorplan is estimated from walking trajectories.}}
    \label{fig:motion_mnist}
\end{figure*}

We show the results of extending \name\ to problems when the forward measurement operator $\mathcal{A}(.)$ is unknown or only partially specified.
We consider the problem of inferring the \emph{spatial map} of an indoor environment (e.g., a home) from a user's walking trajectories from inside that home. 
Floorplans are typically represented as a binary image ($0$ indicating the presence of a wall or a furniture-like object, and $1$ indicating free space), and the human motion model is only partially known --- for instance, walking roughly along the shortest path between two points. 
This naturally lends the problem to blind discrete posterior sampling (recently discussed in \cite{coguide}).
We evaluate on floorplans of resolution $64$ $\times$ $64$ that are obtained from the HouseExpo \cite{houseexpo} dataset.
Walking trajectories are simulated using the $A^*$ algorithm \cite{astar} by choosing random start and end locations. 
At inference time, we pretend the measurement operator is unknown; hence, using the walking trajectories, {\name} must recover the plausible floorplan of the corresponding home.
We report the Intersection over Union (IoU) and F1 score to measure the accuracy of the reconstructed maps.
Table \ref{tab:density_iou_f1} shows that \name\ outperforms the continuous baselines, including CoGuide, on both metrics for the sparse setting in \cite{coguide} and achieves similar performance for the medium-dense case.
Last row in Fig \ref{fig:motion_mnist}. show qualitative results across three different floorplans where we recover ground truth aligned floorplans.

\vspace{-0.5em}
\begin{table}[!htb]
\centering
\scriptsize
\setlength{\tabcolsep}{2pt}
\setlength{\belowcaptionskip}{3pt} 
\begin{tabular}{@{}p{0.56\linewidth}@{\hspace{0.02\linewidth}}p{0.42\linewidth}@{}}
\caption{Spatial mapping performance across trajectory densities in the blind setting. Each regime reports F1-score and IoU metrics.}
\label{tab:density_iou_f1}
&
\caption{Results on music infilling for $\gamma=40\%$ and $\gamma=60\%$. Lower is better for all metrics; best values are in bold. $\Delta$LPS uses $\beta=100$, $T=128$, and $M=40$.}
\label{tab:results_music}
\tabularnewline[-9pt] 
\centering
\resizebox{\linewidth}{!}{%
\begin{tabular}[b]{l *{2}{cc}}
\toprule

\multirow{2}{*}{\textbf{Method}}
  & \multicolumn{2}{c}{\textbf{Sparse}}
  & \multicolumn{2}{c}{\textbf{Moderate}} \\

\cmidrule(lr){2-3} \cmidrule(lr){4-5}

 & \textbf{F1}$\uparrow$ & \textbf{IoU} $\uparrow$
 & \textbf{F1}$\uparrow$ & \textbf{IoU} $\uparrow$ \\

\midrule

\rowcolor{gray!10}
DPS+Neural A$^\star$
  & $0.79 \pm 0.09$ & $0.67 \pm 0.13$
  & $0.79 \pm 0.09$ & $0.66 \pm 0.13$ \\

DPS+TransPath
  & $0.76 \pm 0.15$ & $0.64 \pm 0.18$
  & $0.74 \pm 0.15$ & $0.60 \pm 0.19$ \\

\rowcolor{gray!10}
DPS+DiPPeR
  & $0.77 \pm 0.10$ & $0.64 \pm 0.13$
  & $0.77 \pm 0.11$ & $0.64 \pm 0.14$ \\

DMPlug
  & $0.31 \pm 0.10$ & $0.19 \pm 0.08$
  & $0.28 \pm 0.09$ & $0.17 \pm 0.07$ \\

\rowcolor{gray!10}
DiffPIR
  & $0.63 \pm 0.09$ & $0.47 \pm 0.09$
  & $0.64 \pm 0.08$ & $0.48 \pm 0.09$ \\

CFG
  & $0.86 \pm 0.06$ & $0.76 \pm 0.10$
  & $0.93 \pm 0.03$ & $0.88 \pm 0.05$ \\

\rowcolor{gray!10}
{CoGuide}
  & $0.91 \pm 0.04$ & $0.84 \pm 0.07$
  & $0.94 \pm 0.03$ & $0.89 \pm 0.05$ \\
  
\rowcolor{blue!10}
{\name}
  & $\bm{0.92 \pm 0.06}$ &  $\bm{0.86 \pm 0.09}$
  & $\bm{0.94 \pm 0.03}$ & $\bm{0.89 \pm 0.05}$ \\

\bottomrule
\end{tabular}%
}
&
\centering
\renewcommand{\arraystretch}{1.15} 
\begin{tabular}[b]{lcccc}
\toprule
\textbf{Method}
    & \multicolumn{2}{c}{\textbf{$\gamma=40\%$}}
    & \multicolumn{2}{c}{\textbf{$\gamma=60\%$}} \\

\cmidrule(lr){2-3} \cmidrule(lr){4-5}

 & Hellinger$\downarrow$ & Outlier$\downarrow$
    & Hellinger$\downarrow$ & Outlier$\downarrow$ \\

\midrule


\rowcolor{gray!15}
SVDD-PM
    & 0.465 & 0.112   & 0.529 & 0.114 \\

\rowcolor{gray!15}
SMC
    & 0.492 & 0.152   & 0.552 & 0.164 \\

\rowcolor{gray!15}
SGDD
    & 0.126 & 0.003   & 0.244 & 0.060 \\

\addlinespace[2pt]
\midrule





\rowcolor{blue!10}
$\Delta$\textbf{LPS} 
    & \textbf{0.029} & \textbf{0.000}   & \textbf{0.082} & \textbf{0.000} \\

\bottomrule
\end{tabular}
\tabularnewline
\end{tabular}
\vspace{-1.5em}
\end{table}











\subsection{Monophonic Music Infilling.}
Let us concisely state the task. 
Each music piece is a sequence of length  notes/tokens, where every note is one of  tokens (128 MIDI pitches plus a “silence” token). 
Crucially, the token-to-pitch mapping is scrambled by a fixed random permutation, so that the token indices carry no ordinal meaning. 
As before, we use the pretrained discrete-diffusion prior trained by SGDD and keep it fixed across all methods. 
The forward operator  masks a fraction  {40\%, 60\%} of the notes and reveals the rest. 
Our goal is to infill these missing positions. 
Following SGDD, the likelihood is the exact  measurement model , where counts the mismatches at observed positions and  again controls the guidance strength.
We evaluate our method on 100 test songs and report two metrics as shown in Table \ref{tab:results_music}: the Hellinger distance between the pitch histograms of the generated and ground-truth sequences (lower is better), and the fraction of outliers, i.e. observed positions where the reconstruction disagrees with the measurement, again, lower being better.

\subsection{Guided DNA-Sequence Generation.}
\label{sec:dna}
We consider the task of generating novel DNA sequences following the setup of SGDD \cite{sgdd} and DRAKES \cite{wang2025fine}. 
Each sequence has length $L=200$, and every token is one of the $K=4$ nucleotide bases $A,C,G,T$, making the state space discrete. 
We use SGDD's discrete-diffusion prior and keep it fixed across all our methods.
Unlike the imaging tasks, here the likelihood has no closed-form expression. 
Instead, the goal is to generate sequences with high regulatory activity in the HepG2 cell line, typically scored by a (pretrained) reward model $r(\cdot)$. 
We therefore write the likelihood as $p(\by|\bz_0) \propto \exp(\beta r(\bz_0))$, so the guidance term becomes $\beta \nabla_{\bz_0}r(\bz_0)$. 
Observe that, this is a non-linear task as  is obtained from a neural network. 
The scalar $\beta$ scales the reward gradient and thus controls the guidance strength.
We use the same metrics as in SGDD and report Pred-Activity and ATAC-Acc. 
Pred-Activity is the predicted enhancer activity of the generated sequence under a held-out evaluation oracle (distinct from the reward model used for guidance); we report both its mean and median. 
ATAC-Acc is the fraction of sequences that a pretrained classifier predicts to lie in accessible chromatin, and since active enhancers correlate with chromatin accessibility, this can be used as a proxy for measuring plausible sequences. 
We also report the log likelihood of the generated sequences using the ELBO of a pretrained masked diffusion model [8], which measures how prone a method is to reward-hacking (a sampler could otherwise inflate the reward by drifting toward unrealistic, but high reward sequences).

\vspace{-\ht\strutbox} 
\noindent
\begin{minipage}[t]{0.50\linewidth}
\vspace{0pt}
Table \ref{tab:dna-results} shows the performance of {\name} against other baselines \cite{li2024derivative, wu2023practical, nisonoff2025unlocking, wu2023practical, singhal2025general, wang2025fine, sgdd}, including SGDD. {\name} rows match or exceed SGDD's strongest setting. Note that here T indicates the outer iterations, and M indicates the inner iterations. For instance, at $\beta=7, T = 40, M = 40$, we improve over SGDD across all the metrics: 7.7\% higher reward, 4.8\% higher ATAC-Acc, and a better log-likelihood. Since the likelihood does not degrade, these gains cannot be attributed to reward-hacking. Increasing the sampling budget (inner and outer steps) further helps in improving the performance, up to 9.37 reward with 98.6\% ATAC-Acc. Similar to SGDD, using a stronger guidance scale  improves the reward at the expense of the likelihood, as the sampler moves further away from the prior.
\end{minipage}%
\hfill
\begin{minipage}[t]{0.47\linewidth}
\vspace{0pt}
\centering
\scriptsize
\setlength{\tabcolsep}{2pt}
\setlength{\abovecaptionskip}{0pt}
\setlength{\belowcaptionskip}{3pt}
\captionof{table}{\small{Results on regulatory DNA sequence design. Higher is better for all metrics; best values are in bold. SGDD$_{1}$ and SGDD$_{2}$ use $\beta=30$ and $\beta=50$. {\name}$_{1}$ and {\name}$_{2}$ use $(\beta, T, M) = (7, 40, 40)$ and $(10, 40, 40)$, respectively.}}
\label{tab:dna-results}
\begin{tabular}{lcccc}
\toprule
\textbf{Method}
    & \multicolumn{2}{c}{\textbf{Pred-Activity}}
    & \multirow{2}{*}{\begin{tabular}{@{}c@{}}\textbf{ATAC-}\\\textbf{Acc (\%)$\uparrow$}\end{tabular}}
    & \multirow{2}{*}{\begin{tabular}{@{}c@{}}\textbf{Log-}\\\textbf{likelihood$\uparrow$}\end{tabular}} \\

\cmidrule(lr){2-3}

 & Median$\uparrow$ & Mean$\uparrow$ & & \\

\midrule


\rowcolor{gray!15}
SVDD-PM
    & 5.41 & 5.08 & 49.9 & \textbf{-241} \\

\rowcolor{gray!15}
SMC
    & 4.15 & 3.99 & 39.9 & -259 \\

\rowcolor{gray!15}
CG
    & 2.90 & 2.76 & 0.0 & -265 \\

\rowcolor{gray!15}
TDS
    & 4.64 & 4.62 & 45.3 & -257 \\

\rowcolor{gray!15}
FK-MAX
    & 4.90 & 4.79 & 47.1 & -243 \\

\rowcolor{gray!15}
DRAKES w/o KL
    & 6.44 & 6.28 & 82.5 & -281 \\

\rowcolor{gray!15}
DRAKES w/ KL
    & 5.61 & 5.24 & 92.5 & -264 \\

\rowcolor{gray!15}
SGDD$_{1}$
    & 8.82 & 8.69 & 91.7 & -246 \\

\rowcolor{gray!15}
SGDD$_{2}$
    & 9.14 & 8.96 & 93.0 & -261 \\

\addlinespace[2pt]
\midrule




\rowcolor{blue!10}
{\name}$_{1}$
    & 9.21 & 9.10 & 97.5 & -259.3 \\


\rowcolor{blue!10}
{\name}$_{2}$
    & \textbf{9.37} & \textbf{9.21} & \textbf{98.6} & -268.5 \\

\bottomrule
\end{tabular}
\end{minipage}

\section{Discussion}
\vspace{-0.1in}
We introduced {\name}, a Langevin-inspired posterior sampler for inverse problems in discrete state spaces.
Rather than performing slow sequential Gibbs updates or relying on a continuous relaxation as the final representation, {\name} updates token coordinates in parallel through discrete proposals.
We improved the stability of these proposals by using Adam-preconditioned gradients, which adaptively rescale the likelihood and prior gradients across high-dimensional latent coordinates.

$\blacksquare$ \textbf{Limitations.}
A first limitation is that the method still requires informative gradients of the posterior potential.
For inverse problems with differentiable forward operators, these gradients are directly available, but for highly non-differentiable or black-box measurement processes, one may need surrogate losses, or guided diffusion models as priors.
The quality of the sampler can therefore depend on the fidelity of this gradient signal.
A second limitation is that our current proposal is factorized across token coordinates.
While this enables efficient fully-parallel updates, it may not capture strong dependencies between semantically coupled tokens, especially in highly structured discrete spaces.
Future work could address this through blockwise proposals or via autoregressive correction terms.
Combining {\name} with anchored remasking could help improve the generated posterior samples.
$\blacksquare$ \textbf{Follow-up:} Beyond the tasks shown in this work, an important direction is to study inverse problems over other naturally discrete data.
Many scientific and creative domains are inherently discrete, including biomolecular sequences, symbolic music representations, program tokens, circuit layouts, and structured design spaces.
These settings may benefit more directly from discrete posterior inference, since the target variables are not merely quantized continuous signals but genuinely combinatorial objects.
$\blacksquare$ \textbf{Conclusion.}
As discrete generative models continue to scale to longer sequences, larger vocabularies, and multimodal token spaces, principled inference over discrete representations will become increasingly important.
We view gradient-informed discrete samplers, such as {\name}, as a useful primitive for this broader class of future inverse problems.

\section*{Acknowledgments}
We thank the anonymous reviewers for their valuable feedback.
This work was partially supported by NSF \#2008338, \#1909568, \#2148583, and \#MRI-2018966.
This work used DELTA at NCSA through allocation  CIS230230 from the Advanced Cyberinfrastructure Coordination Ecosystem: Services \& Support (ACCESS) program, which is supported by U.S. National Science Foundation grants \#2138259, \#2138286, \#2138307, \#2137603, \#2551623 and \#2138296.


\bibliography{references}
\bibliographystyle{plain}


\clearpage


\appendix
\clearpage

\etocdepthtag.toc{appendix}
\etocsettagdepth{mainpaper}{none}
\etocsettagdepth{appendix}{subsection}

\etocsettocstyle{%
    \begin{center}
        {\Large\bfseries Appendix Contents}
    \end{center}
    \vspace{0.5em}\hrule\vspace{1.25em}%
}{%
    \vspace{1em}\hrule%
}

\tableofcontents
\clearpage

\section{Algorithm}
Below, we describe our \name{}'s algorithm.
\begin{algorithm}[H]
\caption{\name: Discrete Langevin-Inspired Posterior Sampler}
\label{alg:discrete_langevin_sampler}
\begin{algorithmic}[1]
\Require terminal prior $\pi_T$; measurement $\by$; pretrained discrete diffusion prior $p_\theta(\bz_0 ; \bz_t)$; forward noising kernel $q(\bz_s | \bz_0)$; likelihood $p(\by | \bz_0)$.
\Require diffusion levels $T=t_R>t_{R-1}>\cdots>t_0=0$; inner refinement steps $M$; Langevin step sizes $\{\eta\}$.
\Require Adam parameters $\eta,\gamma_1,\gamma_2,\varepsilon$.
\State $\bz_T \sim \pi_T$ \Comment{\textcolor{gray}{initialize from terminal prior}}
\For{$t=T$ \textbf{down to} $1$}
    \State $r \leftarrow t_r$, \quad $s \leftarrow t_{r-1}$
    \State \textcolor{blue}{$\bz_0^{(0)} \sim p_\theta(\bz_0 | \bz_t)$}
    \Comment{\textcolor{blue}{one-shot denoising using the discrete diffusion prior}}

    \For{$m=0$ \textbf{to} $M-1$}
        \State \textcolor{orange}{$U(\bz_0^{(m)};\bz_t,\by)
        \leftarrow
        \log p(\by | \bz_0^{(m)})
        +
        \log p_\theta(\bz_0^{(m)} | \bz_t)$}

        \State \textcolor{orange}{$\bg^{(m)}
        \leftarrow
        \nabla_{\bz_0} U(\bz_0^{(m)};\bz_t,\by)$}

        \State \textcolor{orange}{$\widetilde{\bg}^{(m)}
        \leftarrow
        \mathrm{Adam}\!\left(\bg^{(m)};\eta_0,\gamma_1,\gamma_2,\varepsilon\right)$}
        \Comment{\textcolor{orange}{Adam-preconditioned gradient}}

        \For{$\ell=1$ \textbf{to} $L$}
            \State \textcolor{teal}{Sample $\bz_0^{(m+1)}[\ell] \sim q_\ell(\cdot |\bz_0^{(m)};\bz_t,\by)$ i.e., Eqn. \ref{eqn:cat_proposal_main}, where}
        \EndFor
    \EndFor
    \State $\bz_0^\star \leftarrow \bz_0^{(M)}$
    \State \textcolor{blue}{$\bz_s \sim p(\bz_s | \bz_0^\star)$}
    \Comment{\textcolor{blue}{re-noise refined sample}}
    \State $\bz_t \leftarrow \bz_s$
\EndFor
\State \Return $\bz_0^\star$
\end{algorithmic}
\end{algorithm}

\section{Additional Quantitative Results}

\label{app:additional_quantitative}

We provide additional results on guided DNA generation and monophonic music infilling, along with ablations on Adam preconditioning and the number of inner refinement steps.
We also compare the runtime and memory usage of \name{}.
Throughout this section, $\beta$ controls the guidance strength, $T$ denotes the number of outer reverse diffusion steps, and $M$ denotes the number of inner refinement steps.
We denote the raw-gradient variant of our method by $\Delta$LPS-G.

\subsection{Guided DNA-Sequence Generation.}

We consider the reward-guided DNA generation task as in the main paper Section \ref{sec:dna} and keep the pretrained discrete diffusion prior fixed across all configurations.
We report the median and mean predicted enhancer activity under a held-out evaluation oracle, ATAC-Acc, and the log-likelihood estimated using the ELBO of a pretrained masked diffusion model.
Table~\ref{tab:app_dna_ablation} reports the results across different guidance strengths and sampling budgets, using both raw gradients ($\Delta$LPS-G) and Adam preconditioning ($\Delta$LPS).

First, for \name{}, increasing the guidance strength from $\beta=5$ to $\beta=7$, with $T=40$ and $M=20$, improves the median predicted activity from $8.48$ to $8.69$ and ATAC-Acc from $89.4\%$ to $91.2\%$.
Increasing the number of inner steps from $M=20$ to $M=40$ at $\beta=7$ further improves the median activity to $9.21$ and ATAC-Acc to $97.5\%$.
Our final configuration, with $\beta=10$, $T=40$, and $M=40$, achieves the highest median and mean activity of $9.37$ and $9.21$, respectively, along with $98.6\%$ ATAC-Acc.
However, perhaps unsurprisingly, stronger guidance also reduces the log-likelihood, indicating a tradeoff between increasing the reward and remaining close to the prior.

Next, we ablate the usefulness of Adam by comparing \name{} with {\name}-G under the same configurations.
Adam consistently improves the median activity by approximately $2.9$--$3.5$ and the mean activity by $3.0$--$3.9$.
The improvements also extend to ATAC-Acc, which increases from $52.7\%$ to $98.6\%$ for the final configuration.
The raw-gradient variant has higher log-likelihoods, but substantially lower activity and ATAC-Acc.
These results support the usefulness of Adam preconditioning in the DNA setting, where guidance is obtained from a nonlinear reward network.

\begin{table}[t]
\centering
\small
\setlength{\tabcolsep}{4pt}
\caption{Guided DNA generation across guidance strengths and sampling budgets, with and without Adam preconditioning.}
\label{tab:app_dna_ablation}
\begin{tabular}{lccccccc}
\toprule
\multirow{2}{*}{\textbf{Method}}
& \multirow{2}{*}{$\beta$}
& \multirow{2}{*}{$T$}
& \multirow{2}{*}{$M$}
& \multicolumn{2}{c}{\textbf{Pred-Activity}}
& \multirow{2}{*}{\textbf{ATAC-Acc (\%)$\uparrow$}}
& \multirow{2}{*}{\textbf{Log-likelihood$\uparrow$}} \\
\cmidrule(lr){5-6}
& & & & Median$\uparrow$ & Mean$\uparrow$ & & \\
\midrule
{\name}-G & 5 & 40 & 20
& 4.95 & 4.48 & 43.9 & -242.9 \\
{\name}-G & 7 & 40 & 20
& 5.77 & 5.33 & 54.4 & -243.9 \\
{\name}-G & 7 & 40 & 40
& 5.70 & 5.35 & 48.1 & -242.0 \\
{\name}-G & 10 & 100 & 20
& 6.20 & 6.06 & 43.9 & \textbf{-228.1} \\
{\name}-G & 10 & 40 & 40
& 6.28 & 6.22 & 52.7 & -240.4 \\
\midrule
{\name} & 5 & 40 & 20
& 8.48 & 8.36 & 89.4 & -248.1 \\
{\name} & 7 & 40 & 20
& 8.69 & 8.58 & 91.2 & -255.6 \\
{\name} & 7 & 40 & 40
& 9.21 & 9.10 & 97.5 & -259.3 \\
{\name} & 10 & 100 & 20
& 9.28 & 9.05 & 93.4 & -260.8 \\
{\name} & 10 & 40 & 40
& \textbf{9.37} & \textbf{9.21} & \textbf{98.6} & -268.5 \\
\bottomrule
\end{tabular}
\vspace{-1em}
\end{table}

\subsection{Monophonic Music Infilling.}

We evaluate music infilling on the same $100$ test songs with masking fractions of $40\%$ and $60\%$.
As in the main paper, we report the Hellinger distance between the generated and ground-truth pitch histograms, and the outlier fraction (measures disagreement with the observed notes).
Table~\ref{tab:app_music_guidance} reports the results across guidance strengths and sampling budgets.

Increasing $\beta$ improves both the Hellinger distance and measurement consistency.
At $T=40$ and $M=50$, increasing $\beta$ from $20$ to $100$ reduces the Hellinger distance from $0.064$ to $0.036$ at $\gamma=40\%$, and from $0.128$ to $0.097$ at $\gamma=60\%$.
The reported outlier fraction is zero at both masking ratios for $\beta\geq50$.
Using $T=128$ and $M=40$ at $\beta=100$ further reduces the Hellinger distances to $0.029$ and $0.082$, respectively, while maintaining zero reported outliers.

\begin{table}[h]
\centering
\small
\setlength{\tabcolsep}{6pt}
\caption{Music infilling across guidance strengths and sampling budgets. $\gamma$ denotes the fraction of masked notes.}
\label{tab:app_music_guidance}
\begin{tabular}{ccccccc}
\toprule
\multirow{2}{*}{$\beta$}
& \multirow{2}{*}{$T$}
& \multirow{2}{*}{$M$}
& \multicolumn{2}{c}{$\gamma=40\%$}
& \multicolumn{2}{c}{$\gamma=60\%$} \\
\cmidrule(lr){4-5}\cmidrule(lr){6-7}
& & & Hellinger$\downarrow$ & Outlier$\downarrow$
& Hellinger$\downarrow$ & Outlier$\downarrow$ \\
\midrule
20 & 40 & 50
& 0.064 & 0.019 & 0.128 & 0.032 \\
50 & 40 & 50
& 0.042 & {0.000} & 0.107 & {0.000} \\
100 & 40 & 50
& 0.036 & {0.000} & 0.097 & {0.000} \\
100 & 128 & 40
& \textbf{0.029} & \textbf{0.000}
& \textbf{0.082} & \textbf{0.000} \\
\bottomrule
\end{tabular}
\vspace{-1em}
\end{table}

Next, Table~\ref{tab:app_music_adam} compares raw gradients with Adam preconditioning.
Interestingly, unlike the DNA task, both variants give essentially identical Hellinger distances across the evaluated configurations.
This is consistent with the music likelihood providing gradients with similar magnitudes at the observed positions, leaving less scope for Adam's coordinate-wise rescaling to improve performance.

\begin{table}[h]
\centering
\small
\setlength{\tabcolsep}{8pt}
\caption{Effect of Adam preconditioning on music infilling across various settings. We compare {\name}-G (raw gradients) and {\name} (Adam preconditioning) under matched configurations.}
\label{tab:app_music_adam}
\begin{tabular}{cccccc}
\toprule
\multirow{2}{*}{$\gamma$}
& \multirow{2}{*}{$\beta$}
& \multirow{2}{*}{$T$}
& \multirow{2}{*}{$M$}
& \multicolumn{2}{c}{\textbf{Hellinger$\downarrow$}} \\
\cmidrule(lr){5-6}
& & & & {\name}-G & {\name} \\
\midrule
$40\%$ & 20  & 40  & 50 & 0.0644 & 0.0633 \\
$40\%$ & 50  & 40  & 50 & 0.0417 & 0.0417 \\
$40\%$ & 100 & 40  & 50 & 0.0362 & 0.0362 \\
$40\%$ & 100 & 128 & 40 & 0.0293 & 0.0293 \\
\midrule
$60\%$ & 20  & 40  & 50 & 0.1284 & 0.1284 \\
$60\%$ & 100 & 128 & 40 & 0.0823 & 0.0823 \\
\bottomrule
\end{tabular}
\end{table}

\subsection{Runtime and Memory Usage.}

We report the end-to-end wall-clock runtime and peak GPU memory usage of SGDD and \name{} on both discrete modalities.
All measurements are obtained on a single $23$~GB GPU, with SGDD evaluated on the same hardware using its publicly released code configuration.
Peak VRAM denotes the maximum GPU memory allocated during generation.
Table~\ref{tab:app_dna_runtime} reports the time required to generate all $640$ DNA sequences at batch size $64$.
\name{} takes $9.7$--$23.9$ minutes across the evaluated configurations, compared to $86.3$--$87.0$ minutes for SGDD, giving approximately $3.6$--$9.0\times$ speedup.
The raw-gradient and Adam variants have nearly identical runtimes and the same reported peak memory usage, indicating little additional overhead from Adam.

\begin{table}[h]
\centering
\small
\setlength{\tabcolsep}{8pt}
\caption{Runtime and memory usage on the DNA task.}
\label{tab:app_dna_runtime}
\begin{tabular}{lccccc}
\toprule
\textbf{Method} & $\beta$ & $T$ & $M$
& \textbf{Time (min)$\downarrow$}
& \textbf{VRAM (GB)$\downarrow$} \\
\midrule
SGDD & 30 & -- & -- & 87.0 & 1.71 \\
SGDD & 50 & -- & -- & 86.3 & 1.71 \\
\midrule
{\name}-G & 5 & 40 & 20 & 9.6 & 2.04 \\
{\name}-G & 7 & 40 & 20 & 9.7 & 2.04 \\
{\name}-G & 7 & 40 & 40 & 18.6 & 2.04 \\
{\name}-G & 10 & 100 & 20 & 24.1 & 2.04 \\
{\name}-G & 10 & 40 & 40 & 18.6 & 2.04 \\
\midrule
{\name} & 5 & 40 & 20 & 9.7 & 2.04 \\
{\name} & 7 & 40 & 20 & 9.8 & 2.04 \\
{\name} & 7 & 40 & 40 & 18.5 & 2.04 \\
{\name} & 10 & 100 & 20 & 23.9 & 2.04 \\
{\name} & 10 & 40 & 40 & 18.4 & 2.04 \\
\bottomrule
\end{tabular}
\vspace{-1em}
\end{table}

Table~\ref{tab:app_music_runtime} reports the time required to infill all $100$ test songs at batch size $50$.
\name{} takes $12.6$ seconds with $T=40$ and $M=50$, and $36.8$ seconds with the best configuration of $T=128$ and $M=40$.
SGDD takes $1382$--$1393$ seconds, giving approximately $38$--$110\times$ speedup at comparable memory usage.
SGDD's music configuration uses $100$ reverse steps, each with $5000$ Metropolis--Hastings steps.
Our method avoids these long inner chains and updates all token positions in parallel.

\begin{table}[h]
\centering
\small
\setlength{\tabcolsep}{8pt}
\caption{Runtime and memory usage for music infilling task.}
\label{tab:app_music_runtime}
\begin{tabular}{lcccc}
\toprule
\textbf{Method} & $T$ & $M$
& \textbf{Time (s)$\downarrow$}
& \textbf{VRAM (GB)$\downarrow$} \\
\midrule
SGDD ($\gamma=40\%$) & 100 & -- & 1393 & 0.85 \\
SGDD ($\gamma=60\%$) & 100 & -- & 1382 & 0.85 \\
\midrule
{\name} & 40 & 50 & 12.6 & 0.88 \\
{\name} & 128 & 40 & 36.8 & 0.88 \\
\bottomrule
\end{tabular}
\vspace{-1em}
\end{table}

\subsection{Effect of Inner Refinement Steps.}

Finally, we ablate the effect of number of inner refinement steps on the random inpainting task.
Table~\ref{tab:app_inner_steps} reports PSNR and LPIPS as $M$ increases from $5$ to $50$.
Increasing $M$ from $5$ to $20$ improves PSNR from $25.90$~dB to $27.29$~dB and reduces LPIPS from $0.2864$ to $0.1729$.
Beyond $20$ steps, PSNR remains nearly unchanged, while LPIPS continues to improve.
Thus, additional refinement steps tend to improve perceptual quality even after the gains in PSNR have largely saturated.

\begin{table}[h]
\centering
\small
\setlength{\tabcolsep}{16pt}
\caption{Effect of the number of inner refinement steps on random inpainting. We report mean $\pm$ standard deviation for PSNR (dB)$\uparrow$ and LPIPS$\downarrow$. Best values are shown in \textbf{bold}.}
\label{tab:app_inner_steps}
\begin{tabular}{ccc}
\toprule
\textbf{Inner steps} ($M$)
& \textbf{PSNR$\uparrow$}
& \textbf{LPIPS$\downarrow$} \\
\midrule
5  & $25.8958 \pm 1.5615$ & $0.2864 \pm 0.0708$ \\
10 & $26.8786 \pm 1.5832$ & $0.2158 \pm 0.0600$ \\
15 & $27.1750 \pm 1.6027$ & $0.1884 \pm 0.0506$ \\
20 & $\boldsymbol{27.2903 \pm 1.5756}$ & $0.1729 \pm 0.0491$ \\
40 & $27.2717 \pm 1.5630$ & $0.1554 \pm 0.0449$ \\
50 & $27.2659 \pm 1.5644$ & $\boldsymbol{0.1528 \pm 0.0441}$ \\
\bottomrule
\end{tabular}
\vspace{-1em}
\end{table}

\section{Additional Qualitative Results}

\begin{figure*}[h]
    \centering

    \begin{minipage}{0.49\textwidth}
        \centering
        \includegraphics[width=\linewidth]{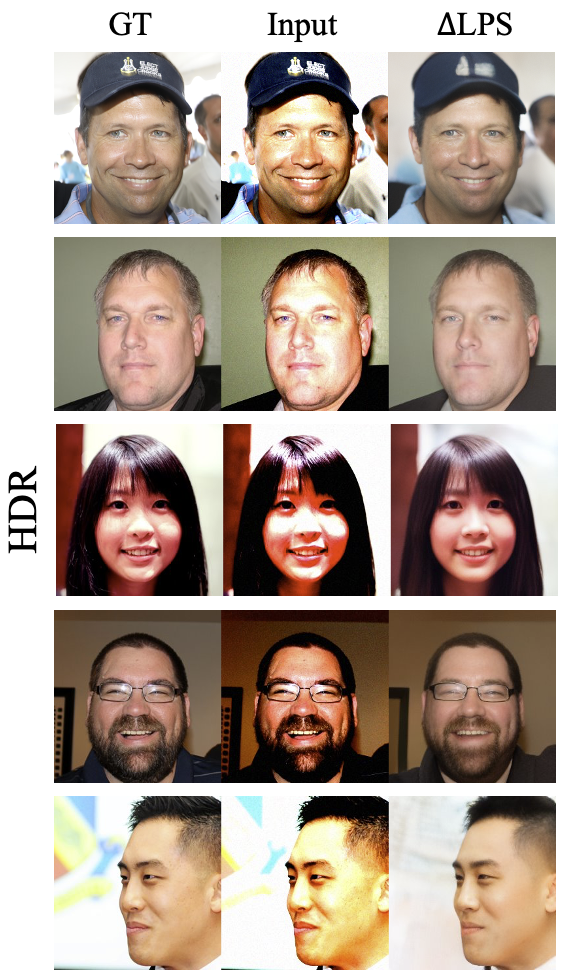}
        \caption{Qualitative results for the HDR reconstruction task on FFHQ. Each example contains the ground truth image, corrupted measurement, and reconstruction from \name{}.}
        \label{fig:ffhq_hdr}
    \end{minipage}
    \hfill
    \begin{minipage}{0.49\textwidth}
        \centering
        \vspace{0.5em}
        \includegraphics[width=\linewidth]{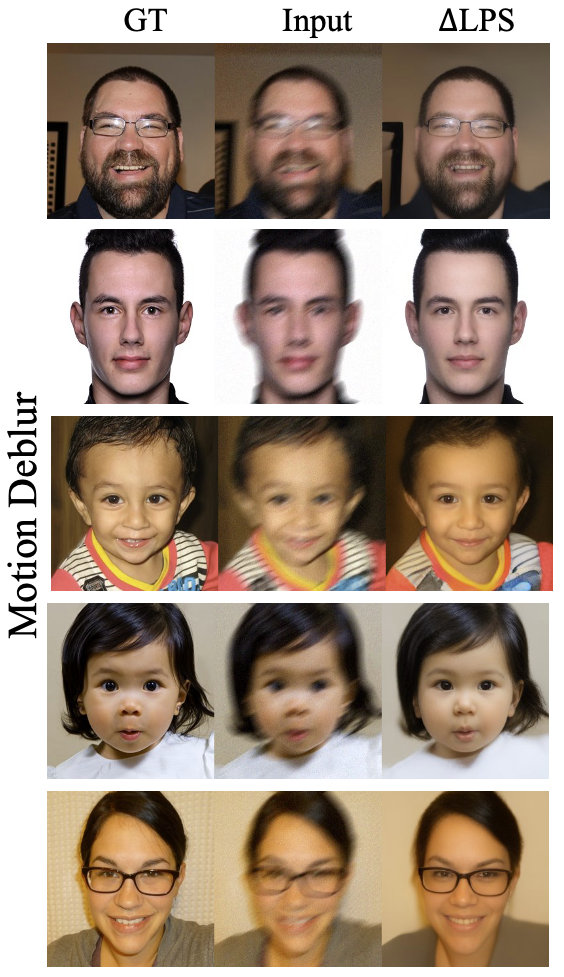}
        \vspace{-0.25em}
        \caption{Qualitative results for motion deblurring on FFHQ. Reconstructions remain sharp and semantically consistent despite severe motion corruption.}
        \label{fig:ffhq_motion}
    \end{minipage}

\end{figure*}

\begin{figure*}[t]
    \centering

    \begin{minipage}{0.49\textwidth}
        \centering
        \includegraphics[width=\linewidth]{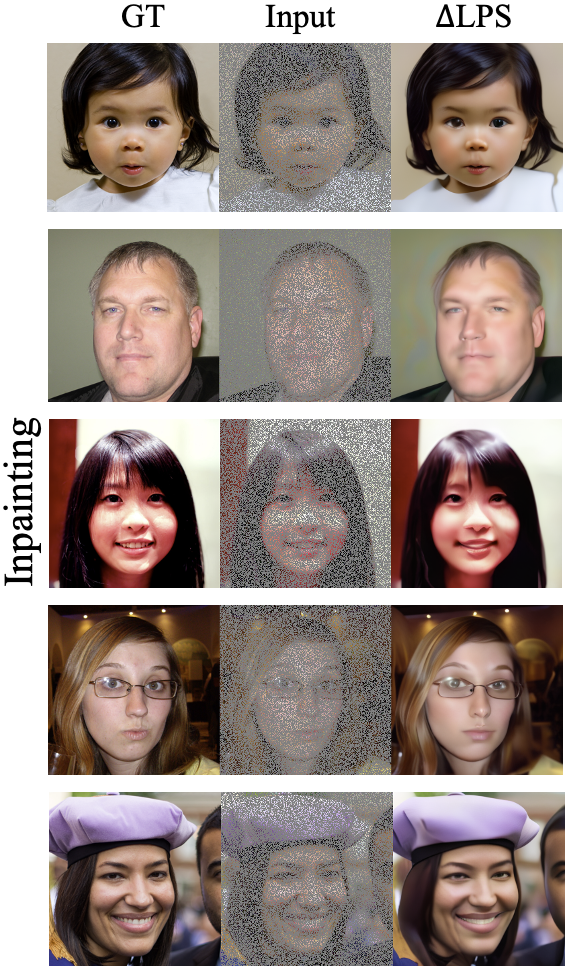}
        \caption{Qualitative results for random inpainting on FFHQ. Even under large missing regions, \name{} generates plausible and visually faithful completions.}
        \label{fig:ffhq_inpaint}
    \end{minipage}
    \hfill
    \begin{minipage}{0.49\textwidth}
        \centering
        \vspace{-1.em}
        \includegraphics[width=\linewidth]{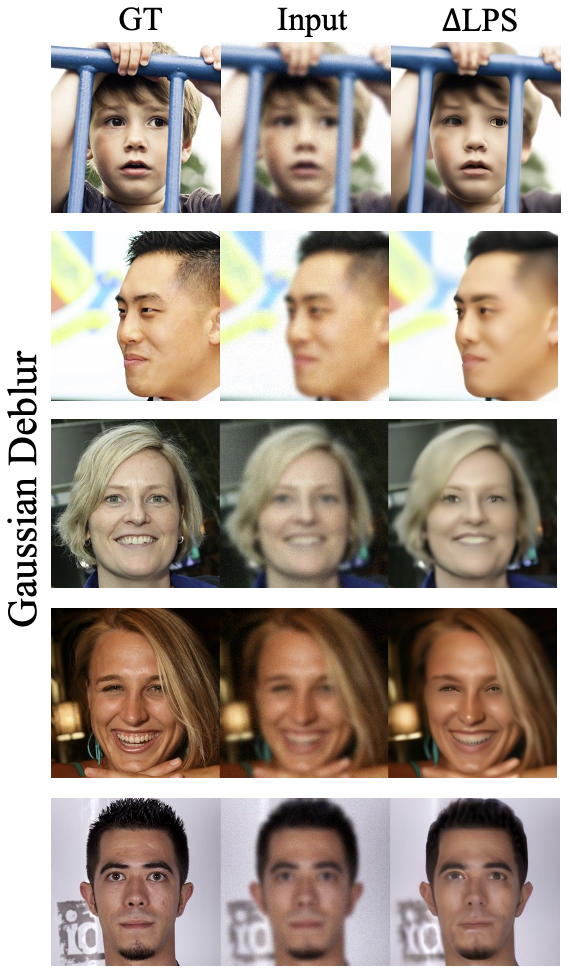}
        \caption{Qualitative results for Gaussian deblurring on FFHQ. \name{} reconstructs realistic facial structures while suppressing blur artifacts.}
        \label{fig:ffhq_deblur}
    \end{minipage}

\end{figure*}

\noindent
We provide additional qualitative reconstructions for FFHQ inverse problems in Figs.~\ref{fig:ffhq_hdr}--\ref{fig:ffhq_inpaint}. 
Each figure shows five representative examples consisting of the ground truth image, the corrupted measurement, and the reconstruction produced by \name{}. 
Across all tasks, \name{} produces visually coherent reconstructions that remain faithful to the original image structure and semantics despite severe corruption. 
In highly ill-posed settings such as HDR reconstruction and motion deblurring, the recovered images preserve realistic facial details and natural textures while remaining consistent with the measurements.


\begin{figure*}[t]
    \centering
    \begin{minipage}{0.49\textwidth}
    \centering
        \includegraphics[width=1\linewidth]{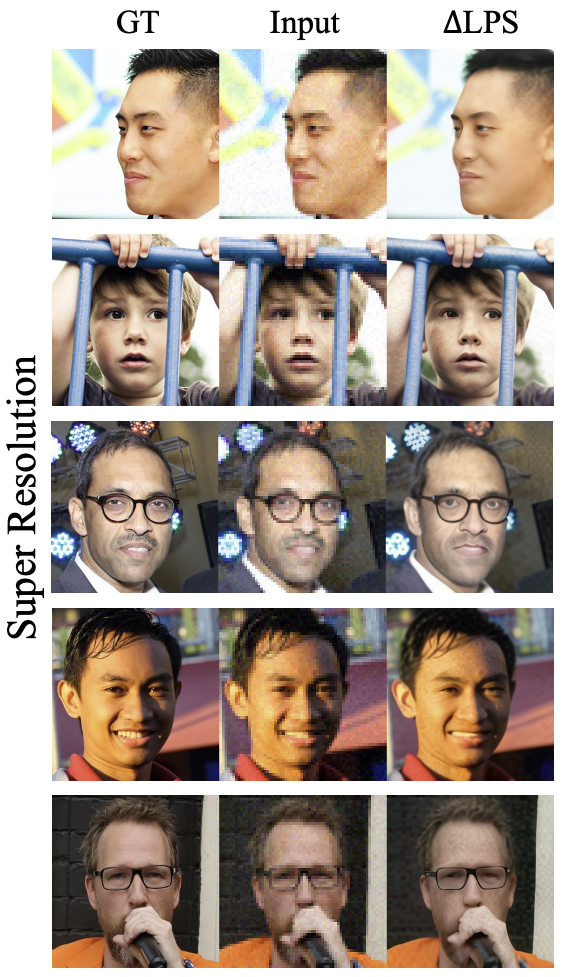}
        \caption{Qualitative results for $4\times$ super-resolution on FFHQ. \name{} recovers fine facial details and realistic textures from low-resolution measurements.}
        \label{fig:ffhq_sr}
    \end{minipage}
    \hfill
    \begin{minipage}{0.49\textwidth}
        \centering
        \vspace{-1.em}
        \includegraphics[width=1\linewidth]{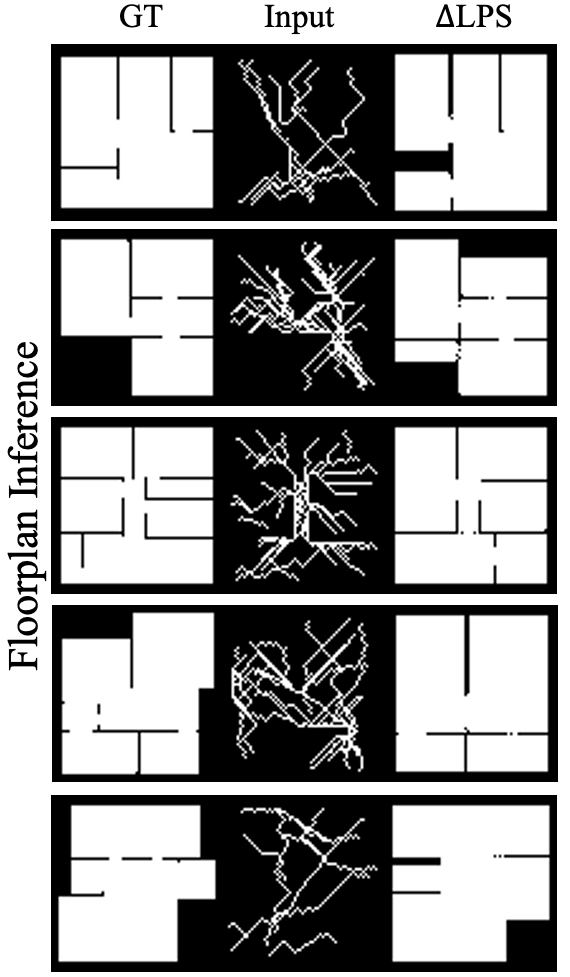}
        \vspace{0.5em}
        \caption{Results for floorplan inference from human walking trajectories. \name{} reconstructs realistic floorplans from sparse measurements.}
        \label{fig:ffhq_deblur}
    \end{minipage}

\end{figure*}

\section{Implementation Platform}

All our experiments are run on 4 NVIDIA A6000 GPUs with 48 GB of VRAM each. All hyperparameters are kept consistent across experiments unless otherwise stated. Random seeds are fixed for reproducibility across runs.

\section{Discrete diffusion models (trained priors)}
\label{app:implementation}

\subsection{Shared backbone architecture (DUO and MDLM)}
\label{app:backbone}

Both trained priors use the same transformer denoiser: a DiT-style discrete diffusion transformer applied to flattened token sequences.
Each token is embedded into a hidden width of $256$.
The stack comprises eight transformer blocks, each with eight attention heads (head dimension $32$), bidirectional self-attention, and rotary positional encodings over the token axis.
Every block applies layer normalization, multi-head attention, dropout at rate $0.1$, and a position-wise MLP expanding the hidden width by a factor of four with GELU nonlinearities.
Noise-level information is passed through a $128$-dimensional embedding that drives AdaLN-style feature modulation.
Under DUO training this embedding tracks the instantaneous forward-process noise coordinate; under MDLM training the coordinate fed to the embedder is identically zeroed so the network is not explicitly conditioned on the current noise magnitude even though the forward process still follows the prescribed schedule.
The output layer applies the same modulation recipe before projecting hidden states to per-token logits.
Representations are additionally scaled by a noise-level-dependent gain (the training configuration enables sigma scaling in the forward pass).

\subsection{DUO (uniform-state discrete diffusion)}
\label{app:duo}

DUO denotes the uniform-state, mean-parameterized discrete diffusion objective trained by ELBO minimization with full noise-level conditioning: the denoiser receives the correct noise coordinate and produces logits that are interpreted as categorical log-probabilities after a softmax.

\subsection{MDLM (masked discrete diffusion)}
\label{app:mdlm}

MDLM denotes the absorbing-state formulation with a substitution-style parameterization for masked positions, likewise trained with an ELBO.
Conditioning on the noise magnitude is disabled as in the preceding subsection.
Ancestral updates use a cached implementation for computational efficiency.
During training, likelihood terms are structured so masked positions carry the learning signal while unmasked positions enforce the instantaneous mask pattern.

\subsection{CIFAR-10 (raw pixel tokens)}
\label{app:cifar}

\paragraph{Data and tokenization.}
Each image becomes a sequence of length $3072$ by rasterizing a $3\times 32\times 32$ tensor in channel-major order.
Every position takes one of $256$ symbol indices corresponding to raw 8-bit channel intensities.

\paragraph{DUO (best CIFAR checkpoint).}
Optimization uses AdamW with zero weight decay, learning rate $2\times 10^{-4}$, $(\beta_1,\beta_2)=(0.9,0.999)$, $\epsilon=10^{-8}$, a $5000$-step linear warmup, and a flat learning rate thereafter.
The global batch size is $96$ across two GPUs, activations use mixed bfloat16 arithmetic with selective float64 stability in sampling, gradients clip at $\ell_2$ norm one, and an exponential moving average of weights uses decay factor $0.9999$.
Training employs antithetic time samples with minimum fractional time $0.01$ when drawing $t$.
The forward process is cosine with cosine noise parameterization and schedule floor $10^{-3}$.
The run targets up to $1.5\times 10^6$ optimizer steps with validation every epoch and model selection on validation negative log-likelihood.
Generation uses $1000$ ancestral updates unless ablated.

\paragraph{MDLM (CIFAR).}
Hyperparameters largely mirror DUO with notable differences: global batch size $128$ on two GPUs; peak learning rate $2.67\times 10^{-4}$; cosine decay after $5000$ warmup steps from $10^{-6}$ down to $10^{-6}$ over $4.95\times 10^5$ decay span inside a $5.0\times 10^5$ step budget; antithetic sampling floor $10^{-3}$; cosine forward process with the same floor; validation and checkpointing every $25$ epochs; monitoring uses $16$ samples every five epochs with $100$ denoising steps.
Sampling swaps in the cached ancestral predictor but otherwise preserves the $1000$-step default.

\subsection{Binary MNIST (padded $32\times 32$)}
\label{app:mnist}

\paragraph{Data and tokenization.}
Images are embedded on a $32\times 32$ canvas (zero padding around $28\times 28$ digits), thresholded at grayscale $128$ to yield binary tokens, and scanned into sequences of length $1024$.
The visible alphabet has two symbols; masked diffusion augments the model vocabulary with a dedicated mask token for absorbing dynamics.

\paragraph{DUO (best MNIST checkpoint).}
Training uses AdamW with zero weight decay, learning rate $3\times 10^{-4}$, default AdamW momentum parameters, $2500$ warmup steps to a constant LR, per-step batch size $128$ on one GPU, mixed bfloat16, gradient clipping at unit norm, and EMA decay $0.9999$.
The noise schedule is log-linear with zero floor on the analytic parameterization and antithetic $t$ draws truncated at $10^{-3}$.
Optimization runs for $10^5$ steps.
Generation defaults to $1000$ ancestral steps with float64 stability where enabled.

\paragraph{MDLM (best MNIST checkpoint).}
Optimization matches DUO's optimizer settings (AdamW, learning rate $3\times 10^{-4}$, $2500$-step warmup, batch $128$, zero weight decay, gradient clip $1.0$, EMA $0.9999$) but trains in full bfloat16 without mixing activations.
Validation fires every $400$ steps; auxiliary checkpoints also save on a fixed step cadence while retaining the lowest-validation model.
The forward process remains log-linear with zero floor; sampling again relies on $1000$ cached ancestral steps with float64 where enabled.

\section{Pretrained multimodal prior and visual tokenizer}
\label{app:mmada_magvit}

For settings that require a large-scale discrete prior without retraining, we adopt the public MMaDA multimodal diffusion language model as a frozen generator.
MMaDA treats text and vision in a unified masked diffusion framework: the backbone initializes from the released LLaDA-$8$B instruct checkpoint, vocabulary is expanded by $8192$-code visual symbols corresponding to $256$ latent image tokens per $256\times 256$ field, and Stage-$1$-style pretraining optimizes with AdamW at learning rate $10^{-4}$, $(\beta_1,\beta_2)=(0.9,0.999)$, weight decay $10^{-2}$, $\epsilon=10^{-8}$, cosine learning-rate decay with $5000$ warmup steps, mixed bfloat16 training, global gradient clipping at unit norm, mild conditioning dropout at probability $0.1$, and on the order of half a million update steps distributed across text-to-image, captioning, and pure language objectives with prescribed loss coefficients and batch mixing rules from the official configuration.

Visual tokens are produced by MagViT-v2, instantiated from the publicly released \emph{Show Lab} MagViT-v2 weights bundled with the MMaDA codebase.
That encoder--decoder maps RGB crops into the $8192$-entry codebook consumed by MMaDA, so image generation, editing, or inversion experiments inherit the $256\times 256$ resolution and tokenizer hyperparameters documented alongside those training recipes.

\section{Inverse problems: forward operators and sampling}
\label{app:inverse_operators}

\subsection{Measurement model and inference backbone}
\label{app:inverse_backbone}

All five tasks assume ground-truth face images in $[0,1]^{3\times 256\times 256}$.
Each inverse problem defines a deterministic forward operator $\mathcal A$ mapping a clean image to an observation $\mathbf y=\mathcal A(\mathbf x)$ in pixel domain, followed by additive Gaussian noise with standard deviation $\sigma=0.05$ applied in the same normalized range convention as each task (linear maps between $[0,1]$ and $[-1,1]$ where the implementation injects noise).
The posterior sampler runs in the discrete latent space of the chosen codec: token logits are pushed through the expected codebook embedding and the frozen decoder so that gradients of the data-fit term reach the categorical variables.
Unless noted otherwise, the data-fit combines weighted $\ell_1$ and squared $\ell_2$ losses on the residual $\mathcal A(\hat{\mathbf x})-\mathbf y$, plus a frozen VGG-$19$ perceptual term on aligned resolutions; discriminator, glue, encoder-observation, and LPIPS terms are disabled in the reported configurations.

Inference uses a DAPS-style discrete posterior sampler with $T$ outer reverse diffusion steps.
At each step the method forms a short-run Markov chain of $K$ discrete Langevin proposals on token space; Metropolis--Hastings correction is turned off (the DULA variant).
Prior scores use one-step posterior logits conditioned on the current noisy tokens.
Inner iterations initialise from the argmax token draw, apply posterior re-noising with unit mixing coefficient, and use a geometric temperature schedule over the $K$ inner steps when $\tau_{\mathrm{start}}\neq\tau_{\mathrm{end}}$.
Proposal distributions use a base inverse-temperature $\alpha$; whenever a strictly smaller floor $\alpha_{\mathrm{min}}$ is supplied, each token uses an entropy-weighted blend between $\alpha_{\mathrm{min}}$ and $\alpha$ so uncertain locations see gentler embedding-space penalties.
Outer-step ratios leave $\alpha$ itself fixed here ($\alpha_{\mathrm{min}}^{\mathrm{ratio}}=1$).
Proposals employ a multiplicative gradient scale on likelihood-informed moves, annealed linearly from an initial value to a terminal value across the outer trajectory when those endpoints differ, and Adam-style accumulation of proposal gradients with $(\beta_1,\beta_2)=(0.9,0.999)$ and stabiliser $\varepsilon=10^{-3}$ without additional Adam beta annealing.
The likelihood weight $\beta$ multiplying the data-fit energy is nominally $\beta_{\max}$ at the end of sampling; whenever a starting value $\beta_0<\beta_{\max}$ is supplied it is linearly ramped from $\beta_0$ at the first outer step to $\beta_{\max}$ at the last (``constant'' schedule refers to the absence of an additional noise-level factor on $\beta$, not to freezing $\beta$ across steps).

For MMaDA codec experiments the prior is text-guided with classifier-free guidance scale $3.5$, maximum text length $512$, and a fixed quality-leaning prompt (natural appearance; sharpened wording for HDR and motion deblur runs matching those JSON presets).
Each configuration draws three stochastic chains per image with the recorded random seed.

\subsection{Gaussian deblurring}
\label{app:op_gaussian_deblur}

The forward operator applies a separable Gaussian blur on $61\times 61$ taps with standard deviation $3$ pixels (implemented as a normalised continuous kernel discretised on that window), with reflect boundary handling, on images mapped to $[-1,1]$ internally before convolution.
This task is evaluated with the MagViT-v2/MM tokenizer stack and MMaDA prior.
Sampling uses $T=30$, $K=30$, $\alpha=0.15$ with floor $\alpha_{\mathrm{min}}=0.01$, unit outer temperatures $\tau$, likelihood ramp $\beta_0=20$ to $\beta_{\max}=80$, proposal gradient scales annealing from $50$ to $20$, and blend coefficients $15$ versus $2$ for the $\mathbf z_0$-anchored reweighting that stabilises early outer steps.
Reconstruction weights are $\lambda_{\ell_1}=3.0$, $\lambda_{\ell_2}=0.5$, $\lambda_{\mathrm{VGG}}=3000$.

\subsection{High dynamic range tone mapping (HDR)}
\label{app:op_hdr}

The forward operator is the pointwise nonlinear $\mathbf y=\mathrm{clip}(2\mathbf x-0.5,\,0,\,1)$, which expands contrast around mid-tones and clips highlights and shadows; it matches the standard ``HDR saturation'' benchmark used alongside linear inverse problems.
Noise is injected after mapping the toned image into $[-1,1]$.
This configuration uses the native FFHQ vector-quantised codec with the same DUO checkpoint as the other FFHQ-codec tasks rather than the MMaDA tokenizer.
Sampling uses $T=30$, $K=30$, matched endpoints $\alpha=\alpha_{\mathrm{min}}=0.2$, inner temperatures cooling geometrically from $\tau_{\mathrm{start}}=2$ to $\tau_{\mathrm{end}}=0.5$, likelihood ramp $1\to 25$, proposal scales $23\to 1$, blend parameters $10$ and $3$, and perceptual weights $\lambda_{\ell_1}=2.2$, $\lambda_{\ell_2}=0.5$, $\lambda_{\mathrm{VGG}}=3000$.

\subsection{Random inpainting}
\label{app:op_inpaint}

A single random pixel mask is drawn per dataset construction with seed tied to the problem instance: approximately $70\%$ of spatial locations are hidden, the same binary pattern is shared across colour channels, retained pixels pass through unchanged in expectation, and missing locations are filled with neutral grey for visualisation.
Gaussian noise applies only on observed coordinates.
This task uses the MMaDA codec path.
Sampling uses $T=20$, $K=20$, $\alpha=0.15$, $\alpha_{\mathrm{min}}=0.01$, uniform $\tau$, likelihood ramp $1\to 25$, constant proposal scaling at $23$, blend $10$ and $3$, and weights $\lambda_{\ell_1}=1.0$, $\lambda_{\ell_2}=0.2$, $\lambda_{\mathrm{VGG}}=3000$.
Only random pixel dropping is considered; spatial box masks are not exercised in this benchmark.

\subsection{$4\times$ super-resolution}
\label{app:op_sr}

Measurements are obtained by a fixed four-fold downsampling of the $256\times 256$ image to $64\times 64$ via a learnable resizer (antialiased bicubic coefficients), producing $\mathbf y$ at low resolution; noise is added in the normalised low-resolution range before returning to $[0,1]$.
Warm-start encodings upsample $\mathbf y$ with bicubic interpolation to full size before tokenisation.
The experiment uses MMaDA.
Sampling sets $T=30$, $K=20$, $\alpha=\alpha_{\mathrm{min}}=0.15$, uniform $\tau$, likelihood ramp $20\to 40$, proposal scales $30\to 5$, blend $10$ and $3$, and $\lambda_{\ell_1}=2.0$, $\lambda_{\ell_2}=0.5$, $\lambda_{\mathrm{VGG}}=3000$.

\subsection{Motion deblurring}
\label{app:op_motion_deblur}

The blur kernel is a $31\times 31$ spatially varying motion pattern generated by the public \emph{motionblur} reference implementation at intensity $0.05$ with deterministic seeding so every instance shares the same kernel; convolution mirrors the Gaussian task (reflect padding, $[-1,1]$ internal dynamics).
Unlike the purely Gaussian PSF, this operator introduces correlated directional smearing typical of camera shake.
The configuration uses the FFHQ codec stack.
Sampling is deeper than for the other tasks: $T=70$ outer steps with $K=30$ inner refinements, large stationary inverse temperatures $\alpha=\alpha_{\mathrm{min}}=1.5$, uniform $\tau$, likelihood ramp $20\to 50.1$, proposal scales $25\to 5$, blend $10$ and $3$, and loss weights $\lambda_{\ell_1}=2.0$, $\lambda_{\ell_2}=0.1$, $\lambda_{\mathrm{VGG}}=5000$ for stronger perceptual emphasis.

\section{From Continuous Langevin Proposals to Discrete Categorical Proposals} \label{app:dlp_derivation}

Our goal is to sample a clean latent sequence $\bz_0$ from a posterior distribution proportional to
$p(\by|\bz_0)p_\theta(\bz_0;\bz_t)$.
Equivalently, we define the log-posterior objective
\begin{align}
    U(\bz_0;\bz_t,\by)
    =
    \log p(\by|\bz_0)
    +
    \log p_\theta(\bz_0;\bz_t).
\end{align}
We restrict the Langevin-inspired proposal to valid discrete candidates and normalize over the finite state space:
\begin{align}
    q(\bz_0'|\bz_0,\bz_t,\by)
    &=
    \frac{
        \exp
        \left(
            -\frac{1}{4\eta}
            \left\|
                \bz_0'
                -
                \bz_0
                -
                \eta
                \nabla_{\bz_0}U(\bz_0;\bz_t,\by)
            \right\|_2^2
        \right)
    }{
        Z(\bz_0)
    }, \label{eqn:dlp_proposal_with_normconst}
\end{align}
where
$Z(\bz_0)
    =
    \sum_{\widetilde{\bz}_0\in\{1,\ldots,K\}^L}
    \exp
    \left(
        -\frac{1}{4\eta}
        \left\|
            \widetilde{\bz}_0
            -
            \bz_0
            -
            \eta
            \nabla_{\bz_0}U(\bz_0;\bz_t,\by)
        \right\|_2^2
    \right)$.
    
Expanding the exponent in Eqn. \ref{eqn:dlp_proposal_with_normconst} gives
\begin{align}
    &-\frac{1}{4\eta}
    \left\|
        \bz_0'
        -
        \bz_0
        -
        \eta
        \nabla_{\bz_0}U(\bz_0;\bz_t,\by)
    \right\|_2^2
    \nonumber
    \\
    &=
    -\frac{1}{4\eta}
    \left\|
        (\bz_0'-\bz_0)
        -
        \eta
        \nabla_{\bz_0}U(\bz_0;\bz_t,\by)
    \right\|_2^2
    \nonumber
    \\
    &=
    -\frac{1}{4\eta}
    \left[
        \|\bz_0'-\bz_0\|_2^2
        -
        2\eta
        \nabla_{\bz_0}U(\bz_0;\bz_t,\by)^\top
        (\bz_0'-\bz_0)
        +
        \eta^2
        \left\|
            \nabla_{\bz_0}U(\bz_0;\bz_t,\by)
        \right\|_2^2
    \right]
    \nonumber
    \\
    &=
    \frac{1}{2}
    \nabla_{\bz_0}U(\bz_0;\bz_t,\by)^\top
    (\bz_0'-\bz_0)
    -
    \frac{1}{4\eta}
    \|\bz_0'-\bz_0\|_2^2
    -
    \frac{\eta}{4}
    \left\|
        \nabla_{\bz_0}U(\bz_0;\bz_t,\by)
    \right\|_2^2.
\end{align}
The term $\left\|\nabla_{\bz_0}U(\bz_0;\bz_t,\by)
    \right\|_2^2$ does not depend on the proposed state $\bz_0'$, so it cancels as the same term appears in the normalization constant. Therefore the overall unnormalized discrete Langevin-inspired proposal can be written as,
\begin{align}
    q(\bz_0'|\bz_0;\bz_t,\by)
    &\propto
    \exp
    \left(
        \frac{1}{2}
        \nabla_{\bz_0}U(\bz_0;\bz_t,\by)^\top
        (\bz_0'-\bz_0)
        -
        \frac{1}{4\eta}
        \|\bz_0'-\bz_0\|_2^2
    \right).
\end{align}

The exponent decomposes across token positions:
\begin{align}
    \frac{1}{2}
    \nabla_{\bz_0}U(\bz_0;\bz_t,\by)^\top
    (\bz_0'-\bz_0)     
    -
    \frac{1}{4\eta}
    \|\bz_0'-\bz_0\|_2^2 \nonumber \\
    &=
    \frac{1}{2}
    \sum_{\ell=1}^L
    \left[
        \nabla_{\bz_0}U(\bz_0;\bz_t,\by)
    \right]_\ell
    \left(
        \bz_0'[\ell]-\bz_0[\ell]\right) \nonumber \\
    &-
     \frac{1}{4\eta}\sum_{\ell=1}^L
    \left(
        \bz_0'[\ell]-\bz_0[\ell]
    \right)^2.
\end{align}
Therefore,
$q(\bz_0'|\bz_0;\bz_t,\by)
    =
    \prod_{\ell=1}^L
    q_\ell(\bz_0'[\ell]|\bz_0,\bz_t,\by),$
where each coordinate has categorical logits
\begin{align}
    r_{\ell,k}
    =
    \frac{1}{2}
    \left[
        \nabla_{\bz_0}U(\bz_0;\bz_t,\by)
    \right]_\ell
    \left(
        k-\bz_0[\ell]
    \right)
    -
    \frac{
        \left(
            k-\bz_0[\ell]
        \right)^2
    }{
        4\eta
    }. \nonumber
\end{align}
Thus, $q_\ell(\bz_0'[\ell]|\bz_0,\bz_t,\by)
    =
    \mathrm{Cat}
    \left(
        \bz_0'[\ell];
        \sigma
        \left(
            r_{\ell,1},\ldots,r_{\ell,K}
        \right)
    \right).$
Equivalently,
\begin{align}
    q(\bz_0'|\bz_0;\bz_t,\by)
    =
    \prod_{\ell=1}^L
    \frac{
        \exp
        \left(
            r_{\ell,\bz_0'[\ell]}
        \right)
    }{
        \sum_{k=1}^K
        \exp
        \left(
            r_{\ell,k}
        \right)
    }.
\end{align}

\subsection{Proposal in Codebook-embedding space}

Now suppose the discrete tokens index a codebook $\mathcal{C}
    =
    \{\bc_1,\ldots,\bc_K\},
    \qquad
    \bc_k\in\R^d.$
The embedding sequence corresponding to $\bz_0$ is $\bc(\bz_0)
    =
    \left[
        \bc_{\bz_0[1]},
        \ldots,
        \bc_{\bz_0[L]}
    \right]
    \in \R^{L\times d}.$
Let
$\bC_0
    =
    \bc(\bz_0),
    \bC_{0,\ell}
    =
    \bc_{\bz_0[\ell]}.$

The posterior objective is
$U(\bz_0;\bz_t,\by)
    =
    \log p(\by|\bz_0)
    +
    \log p_\theta(\bz_0;\bz_t).$
When the measurement likelihood is evaluated through a decoder,
\begin{align}
    \log p(\by|\bz_0)
    =
    \log p
    \left(
        \by
        |
        \oA(\oD(\bc(\bz_0)))
    \right),
\end{align}
we can, in this case, define the differentiable likelihood objective
\begin{align}
    U_y(\bC_0;\by)
    =
    \log p
    \left(
        \by
        |
        \oA(\oD(\bC_0))
    \right).
\end{align}
The diffusion-prior term is defined on discrete tokens:
\begin{align}
    U_\theta(\bz_0;\bz_t)
    =
    \log p_\theta(\bz_0|\bz_t).
\end{align}
Thus,
$U(\bz_0;\bz_t,\by)
    =
    U_y(\bc(\bz_0);\by)
    +
    U_\theta(\bz_0;\bz_t).$

If the full posterior objective were differentiable with respect to the continuous embedding variable $\bC_0$, the Langevin-shaped proposal with variance $2\eta$ would be
\begin{align}
    q(\bC_0'|\bC_0;\bz_t,\by)
    &\propto
    \exp
    \left(
        -\frac{1}{4\eta}
        \left\|
            \bC_0'
            -
            \bC_0
            -
            \eta
            \nabla_{\bC_0}U(\bC_0;\bz_t,\by)
        \right\|_F^2
    \right).
\end{align}
Here, the Frobenius norm may be used instead of the $L_2$ norm. Expanding the exponent gives
\begin{align}
    q(\bC_0'|\bC_0,\bz_t,\by)
    &\propto
    \exp
    \left(
        \frac{1}{2}
            \nabla_{\bC_0}U(\bC_0;\bz_t,\by)^\top
            \bC_0'-\bC_0
        -
        \frac{1}{4\eta}
        \left\|
            \bC_0'-\bC_0
        \right\|_F^2
    \right),
\end{align}
where we have dropped terms independent of the proposal $\bC_0'$.
In our discrete setting, candidate embeddings must come from codebook entries. Therefore, for a proposed token sequence $\bz_0'$, we set
    $\bC_0'
    =
    \bc(\bz_0').$
The likelihood part of the directional derivative is available by backpropagation:
\begin{align}
    \left\langle
        \nabla_{\bC_0}U_y(\bC_0;\by),
        \bc(\bz_0')-\bc(\bz_0)
    \right\rangle.
\end{align}
However, the prior term $U_\theta(\bz_0;\bz_t)$ is defined only on discrete token sequences, so its gradient with respect to $\bC_0$ is not directly available. We approximate its directional contribution by a discrete finite difference over codebook candidates.

For coordinate $\ell$ and candidate token $k$, define the sequence obtained by replacing only the $\ell$-th token of $\bz_0$ with $k$:
\begin{align}
    \bz_0^{(\ell\rightarrow k)}[m]
    =
    \begin{cases}
        k, & m=\ell,\\
        \bz_0[m], & m\neq \ell.
    \end{cases}
\end{align}
The corresponding finite-difference approximation to the prior directional score is
$\Delta_{\theta,\ell,k}
    =
    U_\theta(\bz_0^{(\ell\rightarrow k)};\bz_t)
    -
    U_\theta(\bz_0;\bz_t).$
This approximates
\begin{align}
    \left[
        \nabla_{\bC_{0,\ell}} U_\theta(\bz_0;\bz_t)
    \right]^\top
    \left(
        \bc_k-\bc_{\bz_0[\ell]}
    \right)
    \approx
    \Delta_{\theta,\ell,k}.
\end{align}

Using this finite-difference approximation, the posterior directional score for a full proposal $\bz_0'$ is approximated as
\begin{align}
    &\left\langle
        \nabla_{\bC_0}U(\bC_0;\bz_t,\by),
        \bc(\bz_0')-\bc(\bz_0)
    \right\rangle
    \nonumber
    \\
    &\approx
    \sum_{\ell=1}^L
    \left[
        \nabla_{\bC_{0,\ell}}U_y(\bC_0;\by)
    \right]^\top
    \left(
        \bc_{\bz_0'[\ell]}-\bc_{\bz_0[\ell]}
    \right)
    +
    \sum_{\ell=1}^L
    \Delta_{\theta,\ell,\bz_0'[\ell]}.
\end{align}
Therefore, the codebook-restricted Langevin proposal becomes
\begin{align}
    q(\bz_0'|\bz_0,\bz_t,\by)
    &\propto
    \exp
    \left(
        \sum_{\ell=1}^L
        \frac{1}{2}
        \left[
            \nabla_{\bC_{0,\ell}}U_y(\bC_0;\by)
        \right]^\top
        \left(
            \bc_{\bz_0'[\ell]}-\bc_{\bz_0[\ell]}
        \right)
        \right.
        \nonumber
        \\
        &\qquad\qquad
        \left.
        +
        \sum_{\ell=1}^L
        \frac{1}{2}
        \Delta_{\theta,\ell,\bz_0'[\ell]}
        -
        \sum_{\ell=1}^L
        \frac{1}{4\eta}
        \left\|
            \bc_{\bz_0'[\ell]}-\bc_{\bz_0[\ell]}
        \right\|_2^2
    \right).
\end{align}
Since every term decomposes across token positions, the proposal factorizes:
\begin{align}
    q(\bz_0'|\bz_0,\bz_t,\by)
    =
    \prod_{\ell=1}^L
    q_\ell(\bz_0'[\ell]|\bz_0,\bz_t,\by).
\end{align}

For candidate token $k$ at position $\ell$, the local embedding-space proposal logit is
\begin{align}
    r_{\ell,k}^{\mathrm{emb}}
    =
    \frac{1}{2}
    \left[
        \nabla_{\bC_{0,\ell}}U_y(\bC_0;\by)
    \right]^\top
    \left(
        \bc_k-\bc_{\bz_0[\ell]}
    \right)
    +
    \frac{1}{2}
    \Delta_{\theta,\ell,k}
    -
    \frac{1}{4\eta}
    \left\|
        \bc_k-\bc_{\bz_0[\ell]}
    \right\|_2^2.
\end{align}
Therefore,
\begin{align}
    q_\ell(\bz_0'[\ell]=k|\bz_0,\bz_t,\by)
    =
    \frac{
        \exp
        \left(
            r_{\ell,k}^{\mathrm{emb}}
        \right)
    }{
        \sum_{j=1}^K
        \exp
        \left(
            r_{\ell,j}^{\mathrm{emb}}
        \right)
    }.
\end{align}

If the diffusion prior factorizes across token positions as
\begin{align}
    U_\theta(\bz_0;\bz_t)
    =
    \sum_{\ell=1}^L
    \ell_{\theta,\ell,\bz_0[\ell]}(\bz_t),
\end{align}
then the prior finite difference simplifies to
\begin{align}
    \Delta_{\theta,\ell,k}
    =
    \ell_{\theta,\ell,k}(\bz_t)
    -
    \ell_{\theta,\ell,\bz_0[\ell]}(\bz_t).
\end{align}
Since $\ell_{\theta,\ell,\bz_0[\ell]}(\bz_t)$ is constant with respect to the candidate token $k$, it can be dropped inside the softmax normalization. Hence, an equivalent local logit is
\begin{align}
    r_{\ell,k}^{\mathrm{emb}}
    =
    \frac{1}{2}
    \left[
        \nabla_{\bC_{0,\ell}}U_y(\bC_0;\by)
    \right]^\top
    \left(
        \bc_k-\bc_{\bz_0[\ell]}
    \right)
    +
    \frac{1}{2}
    \ell_{\theta,\ell,k}(\bz_t)
    -
    \frac{1}{4\eta}
    \left\|
        \bc_k-\bc_{\bz_0[\ell]}
    \right\|_2^2.
\end{align}

This was the version that used we in our experiments The likelihood contribution is obtained by a continuous gradient through the decoder, while the prior contribution is approximated by discrete finite differences over candidate codebook embeddings.

\subsection{One-Hot Form}

It is also possible to implement the proposal using one-hot representations of the discrete tokens. Let $\bw_0\in\{0,1\}^{L\times K}$ denote the one-hot representation of $\bz_0$, where $\bw_{0,\ell,k} = \mathbf{1}\{\bz_0[\ell]=k\}.$
Equivalently, each row $\bw_{0,\ell}$ is a one-hot vector
$\bw_{0,\ell} = \be_{\bz_0[\ell]}$,
where $\be_k$ is the $k$-th standard basis vector in $\R^K$.
Let $\ell_{\theta,\ell,k}(\bz_t)$ denote the diffusion-model logit or log-probability assigned to token $k$ at position $\ell$ given $\bz_t$.
A continuous extension of the objective over one-hot variables is
\begin{align}
    U(\bw_0;\bz_t,\by)
    =
    \log p(\by|\bw_0)
    +
    \sum_{\ell=1}^L
    \sum_{k=1}^K
    \bw_{0,\ell,k}
    \ell_{\theta,\ell,k}(\bz_t).
\end{align}
Then the Langevin-shaped proposal over one-hot candidates is
\begin{align}
    q(\bw_0'|\bw_0,\bz_t,\by)
    &\propto
    \exp
    \left(
        -\frac{1}{4\eta}
        \left\|
            \bw_0'
            -
            \bw_0
            -
            \eta
            \nabla_{\bw_0}U(\bw_0;\bz_t,\by)
        \right\|_F^2
    \right).
\end{align}
Expanding and dropping terms independent of $\bw_0'$ gives
\begin{align}
    q(\bw_0'|\bw_0,\bz_t,\by)
    &\propto
    \exp
    \left(
        \frac{1}{2}
        \left\langle
            \nabla_{\bw_0}U(\bw_0;\bz_t,\by),
            \bw_0'-\bw_0
        \right\rangle
        -
        \frac{1}{4\eta}
        \|\bw_0'-\bw_0\|_F^2
    \right). \nonumber
\end{align}
Since both $\bw_0$ and $\bw_0'$ are row-wise one-hot, this again factorizes across token positions:
\begin{align}
    q(\bw_0'|\bw_0,\bz_t,\by)
    =
    \prod_{\ell=1}^L
    q_\ell(\bw_{0,\ell}'|\bw_0,\bz_t,\by). \nonumber
\end{align}
For candidate token $k$, $\bw_{0,\ell}'=\be_k$, so the local logit is
\begin{align}
    r_{\ell,k}^{\mathrm{onehot}}
    =
    \frac{1}{2}
    \left[
        \nabla_{\bw_{0,\ell}}U(\bw_0;\bz_t,\by)
    \right]^\top
    \left(
        \be_k-\bw_{0,\ell}
    \right)
    -
    \frac{1}{4\eta}
    \left\|
        \be_k-\bw_{0,\ell}
    \right\|_2^2. \nonumber
\end{align}
Therefore,
\begin{align}
    q_\ell(\bz_0'[\ell]=k|\bw_0,\bz_t,\by)
    =
    \frac{
        \exp
        \left(
            r_{\ell,k}^{\mathrm{onehot}}
        \right)
    }{
        \sum_{j=1}^K
        \exp
        \left(
            r_{\ell,j}^{\mathrm{onehot}}
        \right)
    }. \nonumber
\end{align}

Since $\bw_{0,\ell}=\be_{\bz_0[\ell]}$, we can also write
\begin{align}
    r_{\ell,k}^{\mathrm{onehot}}
    =
    \frac{1}{2}
    \left[
        \nabla_{\bw_{0,\ell}}U(\bw_0;\bz_t,\by)
    \right]^\top
    \left(
        \be_k-\be_{\bz_0[\ell]}
    \right)
    -
    \frac{1}{4\eta}
    \left\|
        \be_k-\be_{\bz_0[\ell]}
    \right\|_2^2.
\end{align}
Because
\begin{align}
    \left\|
        \be_k-\be_{\bz_0[\ell]}
    \right\|_2^2
    =
    \begin{cases}
        0, & k=\bz_0[\ell], \\
        2, & k\neq \bz_0[\ell],
    \end{cases}
\end{align}
the one-hot form applies a fixed Hamming-distance for changing the token. In particular,
\begin{align}
    r_{\ell,k}^{\mathrm{onehot}}
    =
    \begin{cases}
        0,
        & k=\bz_0[\ell],
        \\
        \frac{1}{2}
        \left[
            \nabla_{\bw_{0,\ell}}U(\bw_0;\bz_t,\by)
        \right]^\top
        \left(
            \be_k-\be_{\bz_0[\ell]}
        \right)
        -
        \frac{1}{2\eta},
        & k\neq \bz_0[\ell].
    \end{cases}
\end{align}

\section{Proof for Contrastive Guidance as a valid surrogate:} \label{app:contra_justification}
For completeness, we include a proof from \cite{coguide} to show the validity of the contrastive surrogate in blind settings. In situations where the likelihood is intractable, {\name} benefits from the replacement of the likelihood term with a contrastive similarity score from the InfoNCE formulation. This relationship was originally shown in Contrastive Predictive Coding \cite{cpcrepresentation}. 
Consider a batch of size $N$ with $\mathcal{\widetilde X}=\{\bx_j\}_{j=1}^N$ and measurements $\mathcal{\widetilde Y}=\{\by_j\}_{j=1}^N$.  
For a given clean data sample $\bx_j$, we shuffle the measurements $\mathcal{\widetilde Y}$, so that the corresponding positive measurement $\by_j$ is placed at position $i\in\{1,\dots,N\}$.  
The task is then to identify the correct index $i$. Let \(I\) be the random variable representing the index of the correct measurement for \(\bx_j\) and \(p(\by)\) be the marginal distribution of measurements.
Given a similarity general score function $s(\bx,\by)$, the InfoNCE assumes a softmax distribution over indices and formulates a cross-entropy loss. So, the approximate posterior \(q\):
\begin{align}
    q(I=i \mid \bx_j,\mathcal{\widetilde Y})
    = \frac{\exp\{s(\bx_j,\by_i)\}}
           {\sum_{k=1}^N \exp\{s(\bx_j,\by_k)\}} .
\end{align}
Specifically, in {\name}, the similarity score function assumes the form $ s(\bx,\by) = \langle f_\varphi(\bx),\; g_\psi(\by) \rangle \,/\, \tau$
since we use encoders $f_{\varphi}$ (for clean data samples), and $g_{\psi}$ (for measurements).
Next, we calculate the true optimal posterior \(p\) from Bayes' rule as :
\begin{align}
p(I=i \mid \bx_j,\mathcal{\widetilde Y})
&= \frac{p(\bx_j,\mathcal{\widetilde Y},I=i)}{\sum_{r=1}^N p(\bx_j,\mathcal{\widetilde Y},I=r)} \nonumber \\ 
&= \frac{\tfrac{1}{N}\,p(\by_i\mid \bx_j)\prod_{k\neq i} p(\by_k)} {\sum_{r=1}^N \tfrac{1}{N}\,p(\by_r\mid x_j)\prod_{k\neq r} p(\by_k)} \nonumber \\ 
&= \frac{p(\by_i\mid \bx_j)\prod_{k\neq i} p(\by_k)} {\sum_{r=1}^N p(\by_r\mid \bx_j)\prod_{k\neq r} p(\by_k)} \nonumber \\ 
&= \frac{\dfrac{p(\by_i\mid \bx_j)}{p(\by_i)}}{\sum_{r=1}^N \dfrac{p(\by_r\mid \bx_j)}{p(\by_r)}} .\nonumber
\end{align}
Matching $q$ and $p$ requires that the numerators differ only by a multiplicative constant (since softmax is invariant to shifts).  
Thus at the InfoNCE optimum,
\begin{align}
    \exp\big(\langle f_\varphi(\bx),g_\psi(\by) \rangle \,/\, \tau\big) &\propto \frac{p(\by|\bx)}{p(\by)} \nonumber \\
    \implies \frac{1}{\tau}\langle f_\varphi(\bx),g_\psi(\by) \rangle &\propto \log\; p(\by| \bx) - \log\; p(\by) \nonumber \\
    \implies \frac{1}{\tau}\langle f_\varphi(\bx),g_\psi(\by) \rangle &= \log\; p(\by| \bx) - \log\; p(\by) + C
\end{align}
where $C$ is independent of $\bx$.  
Taking gradients with respect to $\bx$ cancels both $\log\; p(\by)$ and $C$, yielding
$\frac{1}{\tau}\nabla_x \langle f_\varphi(\bx),g_\psi(\by) \rangle =\nabla_\bx \log\; p(\by| \bx).$


\section{Societal Impact}

{\name} is a general posterior sampling method for inverse problems in discrete state spaces. 
Its potential positive impact lies in enabling more reliable reconstruction and inference for data that are naturally discrete or tokenized. 
This could benefit scientific and engineering applications such as biomolecular design, protein sequence analysis, symbolic music restoration, code completion, circuit/layout design, and compressed media recovery. 
More broadly, improved discrete posterior inference may make generative models more useful for solving constrained reconstruction problems where uncertainty quantification is important.

At the same time, the flexibility of {\name} introduces potential risks. 
Improved inverse solvers can be misused to reconstruct sensitive information from incomplete, corrupted, or indirect measurements, raising privacy concerns in domains such as images, audio, text, biological data, or other tokenized representations. 
Because the method is not tied to a specific prior or measurement operator, it could also be adapted to dual-use settings, including unsafe biological or code-generation objectives. 
Finally, as with other generative-model-based inverse solvers, biases in the pretrained discrete diffusion prior may be preserved or amplified during posterior sampling. 
These concerns motivate careful dataset curation, access controls for sensitive applications, misuse evaluation, and domain-specific safeguards when deploying such methods outside controlled research settings.

\section{Declaration of LLM Usage}
We have used Large Language Models (LLMs) during the preparation of this manuscript to assist with writing, grammar correction, and formatting improvements. The authors reviewed and edited all generated content and take full responsibility for the final version of the manuscript.

\newpage
\clearpage
\section*{NeurIPS Paper Checklist}

The checklist is designed to encourage best practices for responsible machine learning research, addressing issues of reproducibility, transparency, research ethics, and societal impact. Do not remove the checklist: {\bf The papers not including the checklist will be desk rejected.} The checklist should follow the references and follow the (optional) supplemental material.  The checklist does NOT count towards the page
limit. 

Please read the checklist guidelines carefully for information on how to answer these questions. For each question in the checklist:
\begin{itemize}
    \item You should answer \answerYes{}, \answerNo{}, or \answerNA{}.
    \item \answerNA{} means either that the question is Not Applicable for that particular paper or the relevant information is Not Available.
    \item Please provide a short (1--2 sentence) justification right after your answer (even for \answerNA). 
\end{itemize}

{\bf The checklist answers are an integral part of your paper submission.} They are visible to the reviewers, area chairs, senior area chairs, and ethics reviewers. You will also be asked to include it (after eventual revisions) with the final version of your paper, and its final version will be published with the paper.

The reviewers of your paper will be asked to use the checklist as one of the factors in their evaluation. While \answerYes{} is generally preferable to \answerNo{}, it is perfectly acceptable to answer \answerNo{} provided a proper justification is given (e.g., error bars are not reported because it would be too computationally expensive'' or ``we were unable to find the license for the dataset we used''). In general, answering \answerNo{} or \answerNA{} is not grounds for rejection. While the questions are phrased in a binary way, we acknowledge that the true answer is often more nuanced, so please just use your best judgment and write a justification to elaborate. All supporting evidence can appear either in the main paper or the supplemental material, provided in appendix. If you answer \answerYes{} to a question, in the justification please point to the section(s) where related material for the question can be found.

IMPORTANT, please:
\begin{itemize}
    \item {\bf Delete this instruction block, but keep the section heading ``NeurIPS Paper Checklist"},
    \item  {\bf Keep the checklist subsection headings, questions/answers and guidelines below.}
    \item {\bf Do not modify the questions and only use the provided macros for your answers}.
\end{itemize}


\begin{enumerate}

\item {\bf Claims}
    \item[] Question: Do the main claims made in the abstract and introduction accurately reflect the paper's contributions and scope?
    \item[] Answer: \answerYes{} 
    \item[] Justification: Yes, our analysis and experiments support our claims regarding contribution and scope. 

\item {\bf Limitations}
    \item[] Question: Does the paper discuss the limitations of the work performed by the authors?
    \item[] Answer: \answerYes{} 
    \item[] Justification: Yes, we have added a discussion section on the limitations of this work. 

\item {\bf Theory assumptions and proofs}
    \item[] Question: For each theoretical result, does the paper provide the full set of assumptions and a complete (and correct) proof?
    \item[] Answer: \answerNA{} 
    \item[] Justification: This work does not include any mathematical proofs or theorems. 

    \item {\bf Experimental result reproducibility}
    \item[] Question: Does the paper fully disclose all the information needed to reproduce the main experimental results of the paper to the extent that it affects the main claims and/or conclusions of the paper (regardless of whether the code and data are provided or not)?
    \item[] Answer: \answerYes{} 
    \item[] Justification: Yes, we open-source our code to enable reproducibility. Further, our appendix includes details of our models and hyperparameter tunings.

\item {\bf Open access to data and code}
    \item[] Question: Does the paper provide open access to the data and code, with sufficient instructions to faithfully reproduce the main experimental results, as described in supplemental material?
    \item[] Answer: \answerYes{} 
    \item[] Justification: Yes, we have provided enough information to ensure reproducibility.

\item {\bf Experimental setting/details}
    \item[] Question: Does the paper specify all the training and test details (e.g., data splits, hyperparameters, how they were chosen, type of optimizer) necessary to understand the results?
    \item[] Answer: \answerYes{} 
    \item[] Justification: Our appendix includes details of our models and hyperparameter tunings.

\item {\bf Experiment statistical significance}
    \item[] Question: Does the paper report error bars suitably and correctly defined or other appropriate information about the statistical significance of the experiments?
    \item[] Answer: \answerYes{} 
    \item[] Justification: Yes, we have included mean and standard deviation in our experimental section.

\item {\bf Experiments compute resources}
    \item[] Question: For each experiment, does the paper provide sufficient information on the computer resources (type of compute workers, memory, time of execution) needed to reproduce the experiments?
    \item[] Answer: \answerYes{} 
    \item[] Justification: Yes, we have included the details of our compute resources and memory requirements. 
    
\item {\bf Code of ethics}
    \item[] Question: Does the research conducted in the paper conform, in every respect, with the NeurIPS Code of Ethics \url{https://neurips.cc/public/EthicsGuidelines}?
    \item[] Answer: \answerYes{} 
    \item[] Justification: Yes, we have reviewed the NeurIPS code of ethics and our paper conforms to them. 

\item {\bf Broader impacts}
    \item[] Question: Does the paper discuss both potential positive societal impacts and negative societal impacts of the work performed?
    \item[] Answer: \answerYes{} 
    \item[] Justification: We have included this information in the appendix. 
    
\item {\bf Safeguards}
    \item[] Question: Does the paper describe safeguards that have been put in place for responsible release of data or models that have a high risk for misuse (e.g., pre-trained language models, image generators, or scraped datasets)?
    \item[] Answer: \answerNA{} 
    \item[] Justification: Our work poses no such risks.

\item {\bf Licenses for existing assets}
    \item[] Question: Are the creators or original owners of assets (e.g., code, data, models), used in the paper, properly credited and are the license and terms of use explicitly mentioned and properly respected?
    \item[] Answer: \answerYes{} 
    \item[] Justification: Yes, we have credited original owners via citations of their relevant works. 

\item {\bf New assets}
    \item[] Question: Are new assets introduced in the paper well documented and is the documentation provided alongside the assets?
    \item[] Answer: \answerNA{} 
    \item[] Justification: We do not release any new assets. 

\item {\bf Crowdsourcing and research with human subjects}
    \item[] Question: For crowdsourcing experiments and research with human subjects, does the paper include the full text of instructions given to participants and screenshots, if applicable, as well as details about compensation (if any)? 
    \item[] Answer: \answerNA{} 
    \item[] Justification: Our work does not involve crowdsourcing efforts.

\item {\bf Institutional review board (IRB) approvals or equivalent for research with human subjects}
    \item[] Question: Does the paper describe potential risks incurred by study participants, whether such risks were disclosed to the subjects, and whether Institutional Review Board (IRB) approvals (or an equivalent approval/review based on the requirements of your country or institution) were obtained?
    \item[] Answer: \answerNA{} 
    \item[] Justification: Our work does not involve crowdsourcing efforts or participation of human subjects.

\item {\bf Declaration of LLM usage}
    \item[] Question: Does the paper describe the usage of LLMs if it is an important, original, or non-standard component of the core methods in this research? Note that if the LLM is used only for writing, editing, or formatting purposes and does \emph{not} impact the core methodology, scientific rigor, or originality of the research, declaration is not required.
    \item[] Answer: \answerYes{} 
    \item[] Justification: Yes, our appendix includes a section on LLM usage. 

\end{enumerate}

\end{document}